\documentclass[lettersize,journal]{IEEEtran}
\usepackage{amsmath,amsfonts}
\usepackage{algorithmic}
\usepackage{algorithm}
\usepackage{array}
\usepackage[caption=false,font=normalsize,labelfont=sf,textfont=sf]{subfig}
\usepackage{textcomp}
\usepackage{stfloats}
\usepackage{url}
\usepackage{verbatim}
\usepackage{graphicx}
\usepackage{cite}
\usepackage[numbers,sort&compress]{natbib}
\usepackage{color,xcolor}
\usepackage{balance}
\usepackage[misc]{ifsym}
\usepackage{amsthm}
\usepackage{multirow}
\usepackage{booktabs}
\usepackage{hyperref}
\usepackage{makecell}

\usepackage{tikz,xcolor,hyperref}% Make Orcid icon
\definecolor{lime}{HTML}{A6CE39}
\DeclareRobustCommand{\orcidicon}{%
    \begin{tikzpicture}
    \draw[lime, fill=lime] (0,0)
    circle [radius=0.16]
    node[white] {{\fontfamily{qag}\selectfont \tiny ID}};    \draw[white, fill=white] (-0.0625,0.095)
    circle [radius=0.007];    \end{tikzpicture}
    \hspace{-2mm}}
\foreach \x in {A, ..., Z}{%
    \expandafter\xdef\csname orcid\x\endcsname{\noexpand\href{https://orcid.org/\csname orcidauthor\x\endcsname}{\noexpand\orcidicon}}
    }

\begin{document}

\title{A Cross-Scale Hierarchical Transformer with Correspondence-Augmented Attention for inferring Bird’s-Eye-View Semantic Segmentation}

\newcommand{\orcidauthorA}{0000-0003-0145-1690}
\newcommand{\orcidauthorB}{0000-0001-9358-0099}
\newcommand{\orcidauthorD}{0000-0002-5003-3092}
\newcommand{\orcidauthorE}{0000-0002-8055-6468}
\newcommand{\orcidauthorF}{0000-0001-5745-5530}

\author{Naiyu Fang\orcidA{}~\IEEEmembership{Student Member,~IEEE}, Lemiao Qiu\orcidB{}~\IEEEmembership{Member,~IEEE}, Shuyou Zhang, Zili Wang\orcidD{}~\IEEEmembership{Member,~IEEE}, Kerui Hu\orcidE{}, Kang Wang\orcidF{}

\thanks{This paper was produced by the IEEE Publication Technology Group. They are in Piscataway, NJ.}
\thanks{This work was supported by the Natural Science Foundation of Zhejiang Province (LY23E050011), Pioneer and Leading Goose R\&D Program of Zhejiang (2022C01051) and Sichuan Science and Technology Program (2022YFQ0114).}
\thanks{The authors are with State Key Laboratory of Fluid Power \& Mechatronic Systems, Zhejiang University, Hangzhou, 310027, China (e-mail: FangNaiyu@zju.edu.cn; qiulm@zju.edu.cn; zsy@zju.edu.cn; ziliwang@zju.edu.cn; hkr457@zju.edu.cn; wkang@zju.edu.cn.)}
}

\markboth{}
{Shell \MakeLowercase{\textit{Naiyu Fang et al.}}: A Cross-Scale Hierarchical Transformer Infers Bird’s-Eye-View Semantic Segmentation with Correspondence-Augmented Attention}


\maketitle

\begin{abstract}
As bird’s-eye-view (BEV) semantic segmentation is simple-to-visualize and easy-to-handle, it has been applied in autonomous driving to provide the surrounding information to downstream tasks. Inferring BEV semantic segmentation conditioned on multi-camera-view images is a popular scheme in the community as cheap devices and real-time processing. The recent work implemented this task by learning the content and position relationship via the vision Transformer (ViT). However, the quadratic complexity of ViT confines the relationship learning only in the latent layer, leaving the scale gap to impede the representation of fine-grained objects. And their plain fusion method of multi-view features does not conform to the information absorption intention in representing BEV features. To tackle these issues, we propose a novel cross-scale hierarchical Transformer with correspondence-augmented attention for semantic segmentation inference. Specifically, we devise a hierarchical framework to refine the BEV feature representation, where the last size is only half of the final segmentation. To save the computation increase caused by this hierarchical framework, we exploit the cross-scale Transformer to learn feature relationships in a reversed-aligning way, and leverage the residual connection of BEV features to facilitate information transmission between scales. We propose correspondence-augmented attention to distinguish conducive and inconducive correspondences. It is implemented in a simple yet effective way, amplifying attention scores before the Softmax operation, so that the position-view-related and the position-view-disrelated attention scores are highlighted and suppressed. Extensive experiments demonstrate that our method has state-of-the-art performance in inferring BEV semantic segmentation conditioned on multi-camera-view images.
\end{abstract}


\section{Introduction}

\begin{figure}[!ht]
\centering
\includegraphics[width=2.8in]{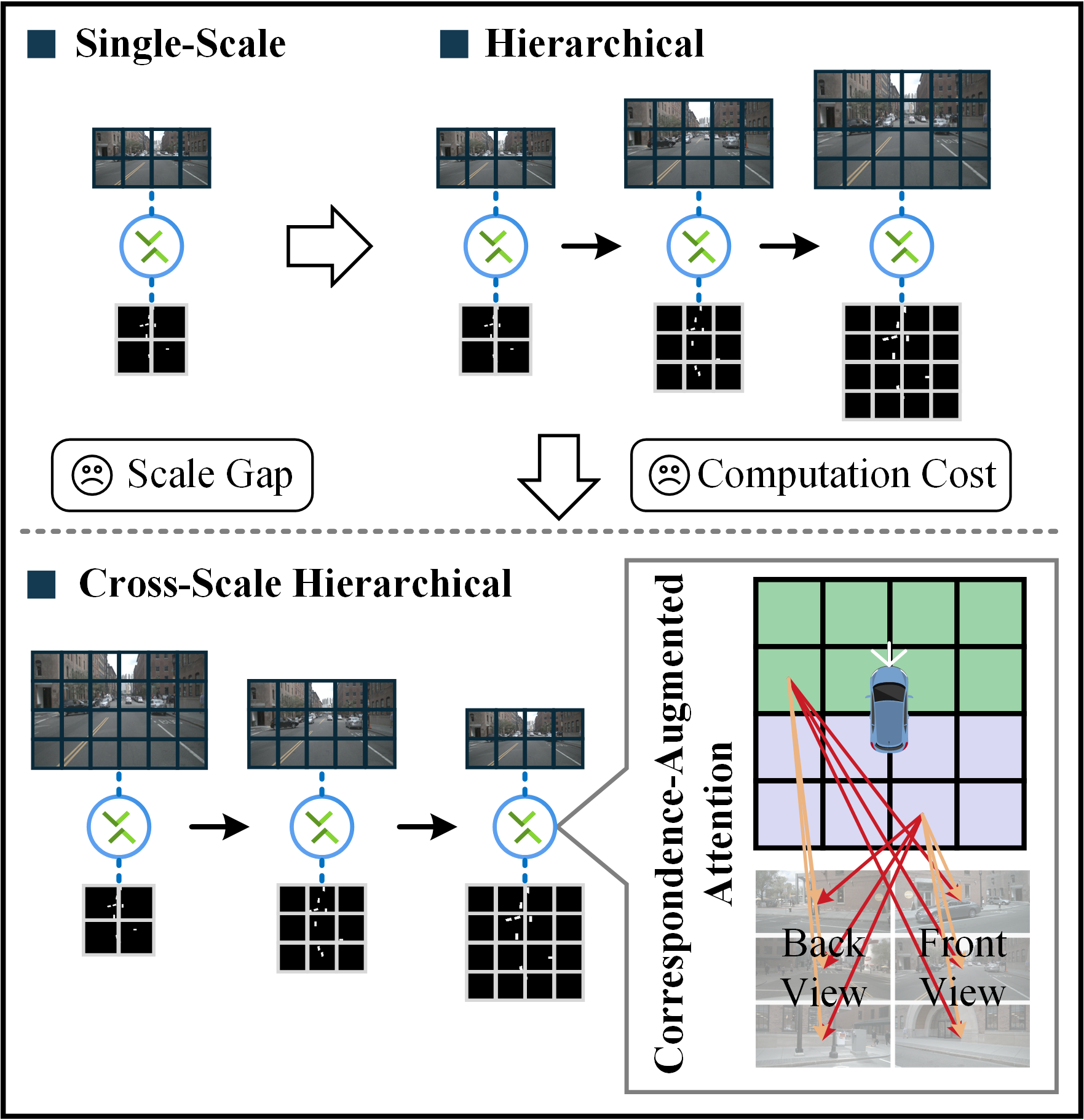}
\caption{The motivation and intention of our proposed method. The cross-scale hierarchical framework is designed to reduce the scale gap between feature interaction and final segmentation without incurring a significant increase in computation. The correspondence-augmented attention aims to highlight the conducive correspondences (red arrows) and suppress inconducive correspondences (yellow arrows).}
\label{fig1}
\end{figure}

\label{sec1}
Perceiving the vehicle surrounding scene is a significant task for autonomous driving, where it provides information on objects and roads for downstream planning, navigation, and steering tasks. As the perceived information form varies from sensor categories, there are two dominant methods to depict surrounding scenes, lidar detection {\color{blue}\citep{li2022modality,li2022panoptic,theodose2021deep,vaquero2020dual}}, and BEV semantic segmentation inference {\color{blue}\citep{zhou2022cross,philion2020lift,rao2023monocular}}. Since BEV semantic segmentation is simple-to-visualize and easy-to-handle, it has been widely utilized and is promising to further bloom autonomous driving. In this paper, towards its hierarchical feature learning and multi-view fusion topics, we endeavor to enable the BEV semantic segmentation inference march toward high-precision and computation-saving.

Camera-view image is the condition in inferring the BEV information. Since they lie in different coordinates but the inductive bias of neural network works in the same coordinate, they {\color{blue}\citep{roddick2020predicting,roddick2018orthographic}} disentangled the BEV semantic segmentation inference task into geometrical relationship conversion and objection detection and classification. They projected pixels or mapped features of camera-view images on the BEV plane and then aggregated or pooled these features to infer BEV information. When it comes to how to establish the geometrical relationship, the earlier works {\color{blue}\citep{ammar2019geometric}} explicitly estimated the projection matrix by calculating the position translation of specific objects. As the considerable coordinate gap, pixel projection is prone to blur and even distort the content of camera-view images on the BEV plane. This downside enables the inferring task merely to trap in pixel warping and object detection {\color{blue}\citep{kim2019deep}}. Drawing inspiration from that, human perceives the surrounding scene in driving without calculating the projection matrix in their brain, neural network can do what human can. The recent works  {\color{blue}\citep{philion2020lift,roddick2020predicting,roddick2018orthographic,rao2023monocular}} learned the implicit geometrical relationship by aggregating the camera-view features into the projected voxel area, which enables the inferring task step into the semantic segmentation. In light of the limited perspective of monocular view, the multi-view information is further fused to enhance the inferring precision and robustness. The semantic Bayesian occupancy grid is leveraged to aggregate multiple predicted BEV maps {\color{blue}\citep{roddick2020predicting}}. This post-processing fusion needs extra computation and enables the model not end-to-end, and the isolated learning for each view impedes the information interaction between views. Consequently, the multi-view fusion is further implemented on the feature level instead of inferred map level {\color{blue}\citep{zhou2022cross}}.

With ViT {\color{blue}\citep{dosovitskiy2021an}} slays in computer vision tasks, the BEV semantic segmentation inference task is comprehended as learning the global relationship between the BEV and camera-view features. This novel task implementation surpasses the traditional feature mapping implementation on inferring precision. However, it also brings new storms to the community where its inherent issues impede its application and promotion. 1) As the quadratic complexity of ViT, the feature relationship learning is restricted in the latent layer, and the size gap between the latent feature and final segmentation impedes the representation of fine-grained objects; 2) The representation of BEV token merely absorbs information from specific not all views. As they fused all view information by concatenation or weighted average, inconducive features will disturb the fusion of conducive features.

To tackle these issues, as shown in {\color{blue} Fig. \ref{fig1}}, we propose a novel method to infer the BEV semantic segmentation, which learns the attention between features from coarse to fine via a cross-scale hierarchical Transformer and reinforces the fusion of conducive features via correspondence-augmented attention. The contributions of our paper are three-fold:

(1) We devise a hierarchical framework to refine the BEV feature representation and diminish the scale gap between the feature and segmentation. The cross-scale Transformer is exploited to save computation, and the residual connection of BEV features is leveraged to facilitate information transmission between scales.

(2) We propose the correspondence-augmented attention to reinforce the relationship between the BEV token and its related view. The attention score is amplified before the Softmax operation to enable the attention scores to distribute toward extreme large and small, so as to highlight conducive correspondences.

(3) Extensive experiments demonstrate that our method has state-of-the-art performance on the challengeable nuScenes dataset {\color{blue}\citep{caesar2020nuscenes}} for both vehicle and road semantic segmentation, compared with inferring BEV semantic segmentation  methods conditioned on multi-camera-view images.

\section{Related Work}
\label{sec2}

\subsection{Surrounding Scene Perceiving}
\label{sec2.1}

The perceiving form of surrounding scenes consists of 2d and 3d forms. Perceiving surrounding scenes in 2d form has the BEV image synthesis and the BEV semantic segmentation inference. The related work of BEV semantic segmentation inference will be described in {\color{blue} Sec. \ref{sec2.2}} in detail. As for the BEV image synthesis, Ammar Abbas S et al. {\color{blue}\citep{ammar2019geometric}} estimated a homography matrix for warping the monitor image into the BEV image. Kim Y et al. {\color{blue}\citep{kim2019deep}} extended this projection and warping into the camera-view image in autonomous driving and exploited a detector to predict vehicle bounding boxes from the BEV image. Instead of estimating an explicit projection matrix, Palazzi A et al. {\color{blue}\citep{palazzi2017learning}} implemented BEV image synthesis by an end-to-end model and supervised model training with the simulated data pair from a game engine. Zhu X et al. {\color{blue}\citep{zhu2018generative}} treated the BEV image synthesis as the image translation and proposed the BridgeGAN to implement the cross-view image translation by cycle consistency. Furthermore, some studies endeavor to capture a maximum of surrounding information from panoramic images by domain shift {\color{blue}\citep{zhang2021transfer}}, contrastive learning {\color{blue}\citep{jaus2023panoramic}}, and optical flow {\color{blue}\citep{shi2023panoflow}}. However, these BEV image synthesis methods are conducted at the pixel level and are prone to producing distortion and artifacts in regions far from the vehicle, which can negatively impact the subsequent detection and segmentation tasks. To enhance the perception precision, we choose to infer BEV semantic segmentation by learning the relationship between image features and BEV features, thus avoiding non-rigid image warping.

Perceiving surrounding scenes in 3d form consists of 3d scene reconstruction and objection detection {\color{blue}\citep{hata2015feature,theodose2021deep,vaquero2020dual}} or segmentation for the point cloud {\color{blue}\citep{peng2022mass,ma2019multi,mei2019semantic}}. Murez Z et al. {\color{blue}\citep{murez2020atlas}} established the scene mesh conditioned on camera-view images, where they leverage 2d CNN to extract 2d image features and back-projected these features into a voxel volume to obtain the scene mesh. Henriques J F et al. {\color{blue}\citep{henriques2018mapnet}} predicted 2.5d scene representation by localizing and registering RGBD inputs. Zhu Z et al. {\color{blue}\citep{zhu2022nice}} incorporated multi-level local information to implement detailed reconstruction in form of a hierarchical feature grid. Object detection and segmentation for the point cloud intends to extract object instances from the raw point cloud. Najibi M et al. {\color{blue}\citep{najibi2022motion}} detected 3d objects and predicted trajectory from LiDAR sequences, where they predicted a motion status to every point and cluster points into semantic concepts. Huang S {\color{blue}\citep{huang2022dynamic}} et al. proposed a spatiotemporal model to segment the static and dynamic objects from multi-frame point clouds. It is worth mentioning that conventional 3D perception methods rely on costly LiDAR sensors. However, with the emergence of a visual-only scheme for autonomous driving, directly perceiving surrounding scenes from camera images becomes a promising approach to lower the device cost.

\subsection{BEV Semantic Segmentation}
\label{sec2.2}
The BEV semantic segmentation inference aims to extract the content features from camera-view images, learn the position relationship between coordinates, and map content features into the BEV plane to infer its semantic segmentation. Early researches utilized monocular camera-view images as the condition to infer the scene at the front of the vehicle. Two crucial technologies boost this research development. Schulter S et al. {\color{blue}\citep{schulter2018learning}} proposed the occlusion-reasoned semantic map to tackle how to represent the BEV semantic segmentation. Roddick T et al. {\color{blue}\citep{roddick2018orthographic}} proposed the orthographic feature transform to tackle how to project camera-view features onto the BEV plane. On this basis, Lu C et al. {\color{blue}\citep{lu2019monocular}} exploited a plain variational encoder-decoder to estimate a BEV semantic segmentation with the size of 64×64 in the cartesian coordinate system. Roddick T et al. {\color{blue}\citep{roddick2020predicting}} regularized the feature projection method, where they extracted the multi-scale monocular features, condensed these features in the vertical axis, and predicted BEV semantic segmentation in the depth axis. Gong S et al. {\color{blue}\citep{gong2022gitnet}} decoupled the BEV map into camera segmentation and geometric prior-based mapping, and leveraged ray-based attention mechanism to aggregate information. Furthermore, Li S et al {\color{blue}\citep{li2023bi}} proposed to incorporate the projection and geometric prior-based mapping for a better performance. These monocular methods have a limited perceiving perspective on the BEV plane. In contrast, we leverage multi-view camera images to broaden the perceiving perspective and improve the perceiving robustness.

Recent works devoted themselves to perceiving the ${360^\circ }$ surrounding scene conditioned on multi-camera-view images. Pan B et al. {\color{blue}\citep{pan2020cross}} concatenated flattened camera-view features to aggregate multi-view information, but their method is implemented on indoor data in simulation environments and is transferred into real environments. Philion J et al. {\color{blue}\citep{philion2020lift}} lifted 2d camera-view features as 3d context vectors by predicting discrete distances, and performed the voxel-wise pooling for multi-view features. Zhou B et al. {\color{blue}\citep{zhou2022cross}} learned the implicit relationship between latent BEV features and multi-scale camera-view features, and embedded the intrinsic and extrinsic matrixes to learn the coordinate transformation. Furthermore, some research proposed spatiotemporal models to further fuse the temporal information. Hu A et al. {\color{blue}\citep{hu2021fiery}} proposed the FIERY to predict non-parametric future trajectories on conditions of past ego-motion information. Li Z et al. {\color{blue}\citep{li2022bevformer}} proposed spatial cross-attention and temporal self-attention to optimize the spatiotemporal fusion in Bevformer. To the best of our knowledge, these methods neglect to explore the correspondence between camera views and BEV tokens, and they allow every BEV token to absorb information from all views, affecting the fusion of conductive features. Therefore, we aim to propose an appropriate correspondence-augmented method to tackle it. Furthermore, they employed relationship learning at a single scale, while we endeavor to establish a hierarchical framework and exploit cross-scale interactions to ensure efficiency.

\begin{figure*}[!ht]
\centering
\includegraphics[width=5.9in]{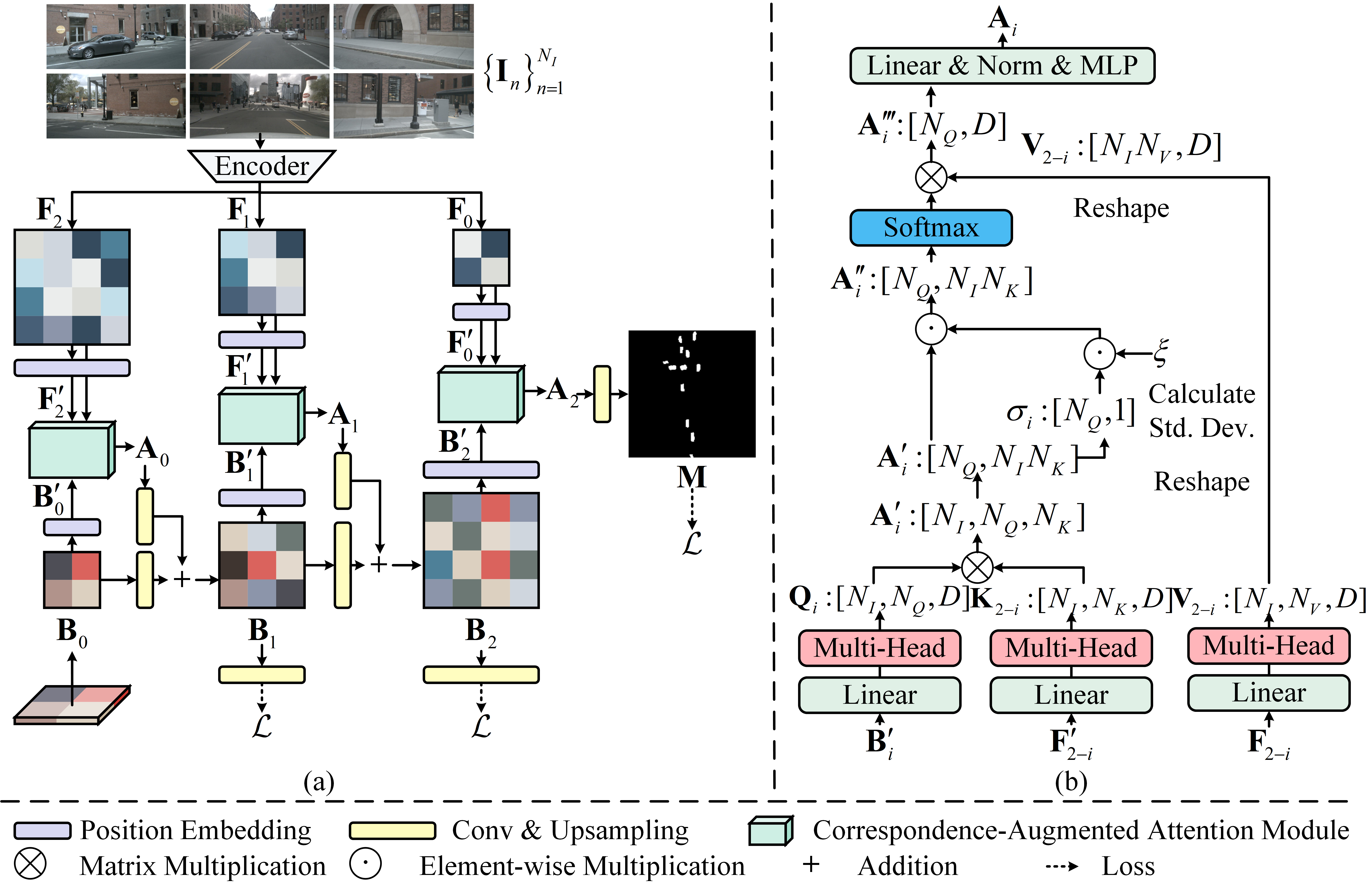}
\caption{(a) The framework of our proposed method. BEV representation learns from ${{{\bf{B}}_{\bf{0}}}}$ to ${{{\bf{B}}_{\bf{2}}}}$. To save computation, they aggregate the cross-scale camera-view features from ${{{\bf{F}}_{\bf{2}}}}$ to ${{{\bf{F}}_0}}$. We then leverage the residual connection of BEV features to facilitate information transmission between scales. And the BEV semantic segmentation ${{\bf{M}}}$ is predicted. (b) The correspondence-augmented attention. It highlights conducive correspondences and suppresses inconducive correspondences by amplifying the attention score before the Softmax function.}
\label{fig2}
\end{figure*}

\section{Methodology}
\label{sec3}
\subsection{Problem Description}
\label{sec3.1}
A vehicle is assumed to have ${{N_I}}$ monocular cameras to form ${360^\circ }$ surrounding perspective. These cameras capture camera-view images and are associated with calibration parameters denoted as ${\left\{ {{{\bf{I}}_n},{{\bf{E}}_n},{{\bf{R}}_n},{{\bf{T}}_n}} \right\}_{n = 1}^{{N_I}}}$, where ${{{\bf{I}}_n} \in {{ \mathbb{R} }^{{H_I} \times {W_I} \times {C_I}}}}$ is a camera-view image with height ${{H_I}}$ and width ${{W_I}}$; ${{{\bf{E}}_n} \in {{ \mathbb{R} }^{3 \times 3}}}$ is the camera intrinsics matrix; ${{{\bf{R}}_n} \in {{ \mathbb{R} }^{3 \times 3}}}$, ${{{\bf{T}}_n} \in {{ \mathbb{R} }^{1 \times 3}}}$ are the camera rotation and translation matrix relative to the vehicle center. We aim to infer the BEV semantic segmentation ${{\bf{M}} \in {{ \mathbb{R} }^{{H_M} \times {W_M} \times {C_M}}}}$ with height ${{H_M}}$ and width ${{W_M}}$ on condition of ${\left\{ {{{\bf{I}}_n},{{\bf{E}}_n},{{\bf{R}}_n},{{\bf{T}}_n}} \right\}_{n = 1}^N}$. To achieve this, a novel and deliberate model is proposed, which learns the attention relationship of feature representation between ${\left\{ {{{\bf{I}}_n}} \right\}_{n = 1}^N}$ and ${{\bf{M}}}$.

\subsection{Cross-Scale Hierarchical Transformer}
\label{sec3.2}
\subsubsection{Hierarchical Framework}
\label{sec3.2.1}

Learning feature correspondence plays a critical role in inferring BEV semantic segmentation. Previous work {\color{blue}\citep{ zhou2022cross}} aggregated multi-scale features of camera-view images into a single-scale BEV feature at the latent layer, and upsampled it to the final size of BEV semantic segmentation. However, this design deviates from our intuition about autonomous driving. Since the granularity at the latent layer is significantly smaller than the final size of BEV semantic segmentation (e.g., 1/8 as set in {\color{blue}\citep{ zhou2022cross}}), it is a challenge to capture distant or small dynamic objects in driving scenes, which enables the downstream planning and steering tasks unstable. To address this issue, as {\color{blue} Fig. \ref{fig2}(a)} shows, we aim to establish a Transformer-based decoder and incorporate the correspondence learning between features into it. Within this hierarchical framework, the BEV feature representation is gradually refined by aggregating the multi-scale features of camera-view images. As a result, the size of the BEV feature at the end of the decoder is only 1/2 of the final size of ${{\bf{M}}}$.

\subsubsection{Cross-Scale Transformer}
\label{sec3.2.2}
We employ a shared encoder ${{\cal E}}$ to extract the features of ${\left\{ {{{\bf{I}}_n}} \right\}_{n = 1}^N}$ at three scales, denoted as ${{{\bf{F}}_0}}$,${{{\bf{F}}_1}}$,${{{\bf{F}}_2}}$. Correspondingly, we set BEV features in three scales as ${{{\bf{B}}_0}}$,${{{\bf{B}}_1}}$,${{{\bf{B}}_2}}$. Since we set size of ${{{\bf{B}}_2}}$ to 1/2 of ${{\bf{M}}}$ and Transformer has the quadratic complexity, in case of the conventional trick, learning the attention score between ${{{\bf{F}}_i}}$ and ${{{\bf{B}}_i}}$ at the corresponding scale will cause tremendous computation, which impacts the real-time autonomous driving. To circumvent this issue, we propose a cross-scale Transformer to align feature scale in a reversed order in this section, namely, BEV feature refines from ${{{\bf{B}}_0}}$ to ${{{\bf{B}}_2}}$, while simultaneously learning the attention relationship with camera-view image features from ${{{\bf{F}}_2}}$ to ${{{\bf{F}}_0}}$. To mitigate the potential inference accuracy loss led by scale misalignment, we introduce a residual connection for each ${{{\bf{B}}_i}}$, facilitating information transmission across scales. We summarize the aforementioned process as, reducing computation by crossing scale and aggregating information by residual connection.

Specifically, as {\color{blue} Fig. \ref{fig2}(a)} shows, the BEV feature ${{{\bf{B}}_i} \in {{ \mathbb{R} }^{H_M^i \times W_M^i \times C_M^i}}}$ grows from a learned ${{{\bf{B}}_0}}$. Following the setting in {\color{blue}\citep{zhou2022cross}}, we embed position information into ${{{\bf{B}}_i}}$ as {\color{blue} Equ. (\ref{eq1})} where ${{{\bf{P}}_i} \in {{ \mathbb{R} }^{H_M^i \times W_M^i \times 2}}}$ is the pixel position matrix of ${{{\bf{B}}_i}}$ in the vehicle coordinate; ${{f_P}\left(  \cdot  \right)}$ and ${{f_T}\left(  \cdot  \right)}$ project ${{{\bf{P}}_i}}$ and ${{\bf{T}}}$ to be the same dimension of ${{{\bf{B}}_i}}$. Similarly, the cross-scale feature ${{{\bf{F}}_{2 - i}}}$ is also embedded with the position information as {\color{blue} Equ. (\ref{eq2})} where ${{{\bf{P}}_{2 - i}} \in {{ \mathbb{R} }^{H_I^{2 - i} \times W_I^{2 - i} \times 2}}}$ is the pixel position matrix of ${{{\bf{F}}_{2 - i}}}$ in the camera-view coordinate, and this position information is transformed into the one in the vehicle coordinate by multiplying with ${{{\bf{E}}^{ - 1}}}$ and ${{{\bf{R}}^{ - 1}}}$; ${{f_F}\left(  \cdot  \right)}$ and ${{f_P}\left(  \cdot  \right)}$ project ${{{\bf{F}}_{2 - i}}}$ and ${{{\bf{R}}^{ - 1}} \cdot {{\bf{E}}^{ - 1}} \cdot {{\bf{P}}_{2 - i}}}$ into the dimension ${H_I^{2 - i} \times W_I^{2 - i} \times C_I^{2 - i}}$.

\begin{equation}
\label{eq1}
{{\bf{B'}}_i} = {{\bf{B}}_i} + {f_P}\left( {{{\bf{P}}_i}} \right) - {f_T}\left( {\bf{T}} \right)
\end{equation}

\begin{equation}
\label{eq2}
{{\bf{F'}}_{2 - i}} = {f_F}\left( {{{\bf{F}}_{2 - i}}} \right) + {f_P}\left( {{{\bf{R}}^{ - 1}} \cdot {{\bf{K}}^{ - 1}} \cdot {{\bf{P}}_{2 - i}}} \right) - {f_T}\left( {\bf{T}} \right)
\end{equation}

We intend to learn the global context relationship between ${{{\bf{B'}}_i}}$ and ${{{\bf{F'}}_{2 - i}}}$ to aggregate the information of ${{{\bf{F'}}_{2 - i}}}$ into ${{{\bf{B'}}_i}}$. ${{{\bf{B'}}_i}}$, ${{{\bf{F'}}_{2 - i}}}$, and ${{{\bf{F}}_{2 - i}}}$ are regarded as query, key, and value. The correspondence-augmented attention module is conditioned on them and computes ${{{\bf{A}}_i}}$. We leverage two \{Conv+Upsample\} layer combinations to project ${{{\bf{A}}_i}}$ and ${{{\bf{B}}_i}}$ into the next scale, and add the projected results together for the residual connection. Ultimately, ${{\bf{M}}}$ is inferred from ${{{\bf{A}}_2}}$ at the last scale.

\subsubsection{Model Complexity}
\label{sec3.2.3}
We further theoretically analyze the computation saving of cross-scale Transformer for the overall hierarchical framework compared to the traditional one. As described in {\color{blue}\citep{han2021transformer}}, the FLOPs of standard multi-head attention (MA) is shown in {\color{blue} Equ. (\ref{eq3})} where ${{N_Q}}$, ${{N_K}}$, ${{N_V}}$ are the token number of query, key, value; ${{D_Q}}$, ${{D_K}}$, ${{D_V}}$ are the token length of query, key, value.

\begin{equation}
\label{eq3}
FLO{P_{MA}} = {N_Q}{N_K}\left( {{D_Q} + {D_k}} \right) + 2{N_Q}D_Q^2 + {N_K}D_K^2 + {N_V}D_V^2
\end{equation}

In our cross-scale Transformer, we simplify the MA case with the conditions of ${{N_K} = {N_V}}$ and ${D = {D_Q} = {D_K} = {D_v}}$. Therefore, the MA FLOPs of this case is calculated as {\color{blue} Equ. (\ref{eq4})}.

\begin{equation}
\label{eq4}
FLO{P_{MA}} = 2{N_Q}{N_K}D + 2{D^2}\left( {{N_Q} + {N_K}} \right)
\end{equation}

We assume three scale groups for our and the traditional hierarchical framework as ${\left\{ {\left( {{N_Q},{N_K}} \right)|} \right.}$${\left( {\frac{1}{8}{H_M}{W_M},\frac{1}{2}{H_I}{W_I}} \right);}$${\left( {\frac{1}{4}{H_M}{W_M},\frac{1}{4}{H_I}{W_I}} \right);}$ ${\left. {\left( {\frac{1}{2}{H_M}{W_M},\frac{1}{8}{H_I}{W_I}} \right)} \right\}}$ and ${\left\{ {\left( {{N_Q},{N_K}} \right)|} \right.}$${\left( {\frac{1}{8}{H_M}{W_M},\frac{1}{8}{H_I}{W_I}} \right);}$ ${\left( {\frac{1}{4}{H_M}{W_M},\frac{1}{4}{H_I}{W_I}} \right);}$${\left. {\left( {\frac{1}{2}{H_M}{W_M},\frac{1}{2}{H_I}{W_I}} \right)} \right\}}$. Thus, their FLOPs are ${\frac{3}{8}{H_M}{W_M}{H_I}{W_I}D + 2{D^2}\left( {{H_M}{W_M} + {H_I}{W_I}} \right)}$ and ${\frac{{21}}{{32}}{H_M}{W_M}{H_I}{W_I}D + 2{D^2}\left( {{H_M}{W_M} + {H_I}{W_I}} \right)}$. Since their latter items are the same and ${D \ll {N_Q}||{N_K}}$ in our setting, the cross-scale Transformer is capable of saving 42.86\% computation for the hierarchical framework.

\subsection{Correspondence-Augmented Attention}
\label{sec3.3}

\subsubsection{Multi-View Fusion Insight}
\label{sec3.3.1}

\begin{figure}[!ht]
\centering
\includegraphics[width=3.3in]{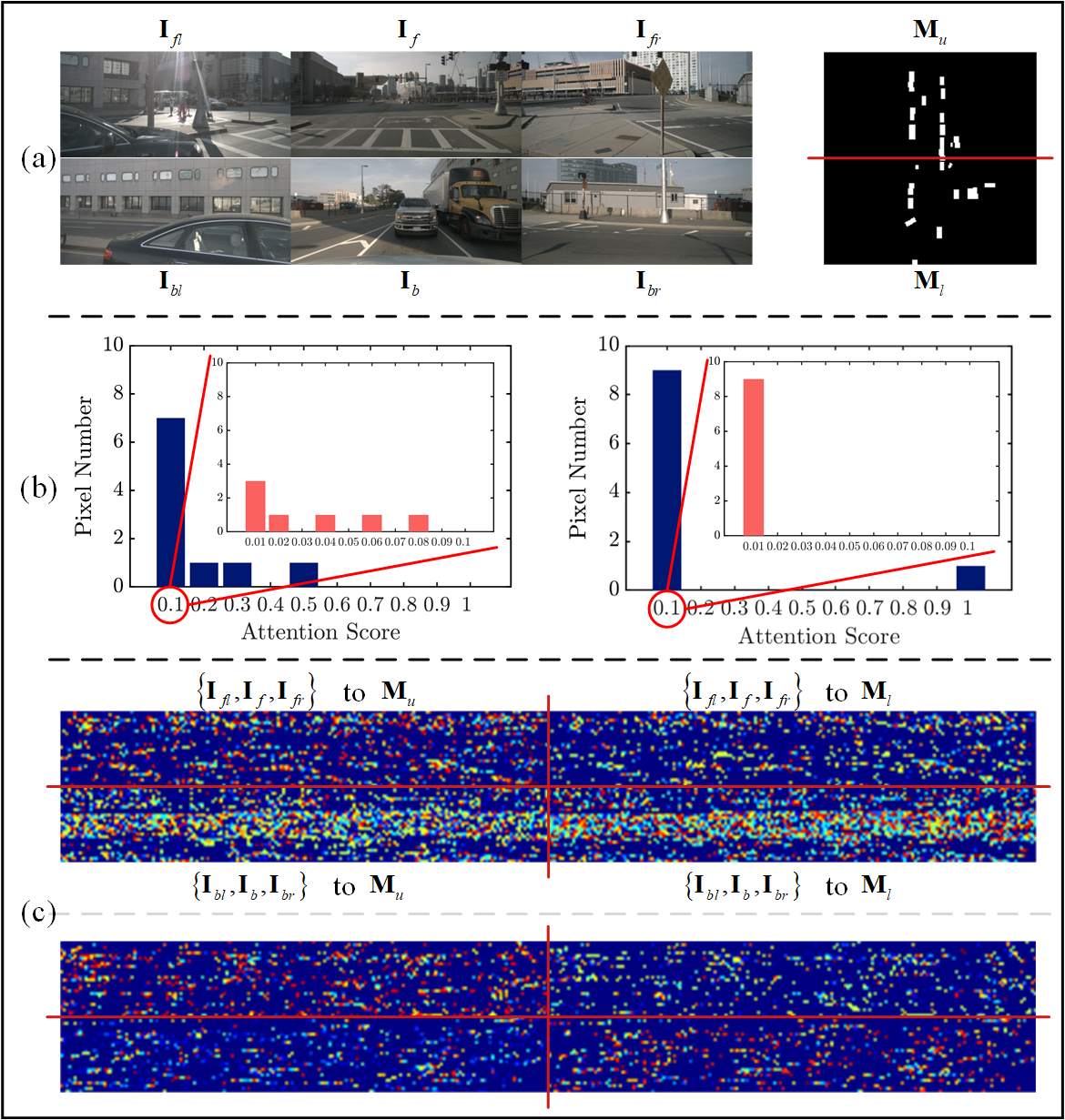}
\caption{(a) The symbols of camera-view images and BEV semantic segmentation; (b) The attention score variation in correspondence augmentation, the left and right parts are without and with attention score amplification before softmax operation; (c) The correspondence between camera-view images and BEV semantic segmentation, the upper and lower parts are without and with correspondence augmentation.}
\label{fig3}
\end{figure}

Intuitively, since every camera has a limited perspective and these ${{N_I}}$ cameras have different positions and orientations, the learning of representing ${{\bf{M}}}$ is region-wise. A single camera-view image provides valuable information only to a specific region of ${{\bf{M}}}$, consequently, most regions in ${{\bf{M}}}$ do not require information from all camera views. For instance, as shown in {\color{blue} Fig. \ref{fig3}(a)}, ${{{\bf{I}}_b}}$ contains perspective information behind the vehicle and should primarily contribute to the representation of the lower part of ${{\bf{M}}}$ denoted as ${{{\bf{M}}_l}}$. Similarly, ${{{\bf{M}}_u}}$ represents object cases in front of the vehicle and should utilize information from ${\left\{ {{{\bf{I}}_{fl}},{{\bf{I}}_f},{{\bf{I}}_{fr}}} \right\}}$ instead of ${\left\{ {{{\bf{I}}_{bl}},{{\bf{I}}_b},{{\bf{I}}_{br}}} \right\}}$.

The previous work {\color{blue}\citep{ zhou2022cross}} employed the Transformer to learn the attention score of a BEV token with all camera-view tokens for all views. These attention scores were then used to aggregate the features of all views through a weighted average, resulting in each BEV token fusing information from all camera-view tokens across all views. It seems that this thing is good, not so! The upper part of {\color{blue} Fig. \ref{fig3}(c)} illustrates the correspondence of each camera-view token to either ${{{\bf{M}}_u}}$ or ${{{\bf{M}}_l}}$, with correspondence referring to the number of attention scores greater than a threshold. Surprisingly, it shows that both ${\left\{ {{{\bf{I}}_{fl}},{{\bf{I}}_f},{{\bf{I}}_{fr}}} \right\}}$and ${\left\{ {{{\bf{I}}_{bl}},{{\bf{I}}_b},{{\bf{I}}_{br}}} \right\}}$ have similar levels of correspondence to ${{{\bf{M}}_l}}$. As attention scores are normalized using the Softmax function, the correspondence from ${\left\{ {{{\bf{I}}_{fl}},{{\bf{I}}_f},{{\bf{I}}_{fr}}} \right\}}$ reduces the attention score of ${\left\{ {{{\bf{I}}_{bl}},{{\bf{I}}_b},{{\bf{I}}_{br}}} \right\}}$ when representing ${{{\bf{M}}_l}}$. In other words, inconducive features disturb the fusion of conducive features, which is inconsistent with our intuition. In contrast, we aim to emphasize conducive correspondences while suppressing inconducive correspondences, specifically by increasing position-view-related attention scores and decreasing position-view-disrelated attention scores.

\subsubsection{Correspondence Augmentation}
\label{sec3.3.2}
Although the conducive and inconducive correspondences have the same order of magnitude, a well-trained model should assign a larger position-view-related attention score than the position-view-disrelated one. Based on this assumption, we classify large and small attention scores as conducive and inconducive correspondences, respectively. A straightforward approach would be to reserve only the maximum or top-k attention scores and aggregate their corresponding camera-view features. However, this would lead to a non-differentiable operation in discarded parts and result in unstable model training. To address this, instead of discarding attention scores, we aim to encourage large attention scores to be close to 1 and smaller ones to be close to 0. As shown in {\color{blue} Fig. \ref{fig3}(b)}, amplifying overall attention scores before the Softmax normalization causes them to distribute toward extremes, aligning with our expected correspondence distribution. Drawing inspiration from this, we propose a simple yet effective correspondence-augmented attention mechanism to address the issue described in {\color{blue} Sec. \ref{sec3.3.1}}.

To be specific, for each view, we first learn its attention scores with the BEV feature. ${{{\bf{B'}}_i}}$, ${{{\bf{F'}}_{2 - i}}}$, ${{{\bf{F}}_{2 - i}}}$ are considered as ${{{\bf{Q}}_i} \in {{ \mathbb{R} }^{{N_I} \times {N_Q} \times D}}}$, ${{{\bf{K}}_{2 - i}} \in {{ \mathbb{R} }^{{N_I} \times {N_K} \times D}}}$, ${{{\bf{V}}_{2 - i}} \in {{ \mathbb{R} }^{{N_I} \times {N_V} \times D}}}$ through a linear layer and the multi-head transition, and the attention scores of all views are concatenated to obtain ${{{\bf{A'}}_i} \in {{ \mathbb{R} }^{{N_I} \times {N_Q} \times {N_K}}}}$. Subsequently, we introduce an amplification step to modify the distribution of ${{{\bf{A'}}_i}}$. As the attention score distribution varies with the position of BEV token, we compute a local amplification coefficient for each BEV token. ${{{\bf{A'}}_i}}$ is reshaped as the size ${{N_Q} \times {N_I}{N_K}}$, a standard deviation ${{\sigma _i} \in {{ \mathbb{R} }^{{N_Q} \times 1}}}$ is calculated along the dimension ${{N_I}{N_K}}$, and ${{\sigma _i}}$ is multiplied with a coefficient ${\xi }$ to determine the augmentation degree. We calculate the augmented attention score ${{{\bf{A''}}_i} \in {{ \mathbb{R} }^{{N_Q} \times {N_I}{N_K}}}}$ as {\color{blue} Equ. (\ref{eq5})} and aggregate the multi-view image features to yield ${{{\bf{A'''}}_i}}$. Finally, ${{{\bf{A}}_i} \in {{ \mathbb{R} }^{H_M^i \times W_M^i \times C_M^i}}}$ is derived through the traditional combination of linear layer, normalization, and MLP.

\begin{equation}
\label{eq5}
{{\bf{A''}}_i} = \xi  \cdot {\sigma _i} \cdot {{\bf{A'}}_i}
\end{equation}

As shown in the lower part of {\color{blue} Fig. \ref{fig3}(c)}, the correspondence-augmented attention enables the correspondence ${\left\{ {{{\bf{I}}_{fl}},{{\bf{I}}_f},{{\bf{I}}_{fr}}} \right\}}$ to ${{{\bf{M}}_u}}$ and ${\left\{ {{{\bf{I}}_{bl}},{{\bf{I}}_b},{{\bf{I}}_{br}}} \right\}}$ to ${{{\bf{M}}_l}}$ much larger than ${\left\{ {{{\bf{I}}_{fl}},{{\bf{I}}_f},{{\bf{I}}_{fr}}} \right\}}$ to ${{{\bf{M}}_l}}$ and ${\left\{ {{{\bf{I}}_{bl}},{{\bf{I}}_b},{{\bf{I}}_{br}}} \right\}}$ to ${{{\bf{M}}_u}}$, which conforms to our expectation described in {\color{blue} Sec. \ref{sec3.3.1}}.

\subsection{Objective Function}
\label{sec3.4}
We supervise the learning of BEV representation from the initial learned feature to the final output for the cross-scale hierarchical Transformer. Its loss is defined as {\color{blue} Equ. (\ref{eq6})} where ${F\left( { \cdot , \cdot } \right)}$ is focal loss {\color{blue}\citep{ lin2017focal}}; ${f\left(  \cdot  \right)}$ is the layer combination of convolution and upsampling, which projects ${{{\bf{B}}_i}}$ into the size of ${{H_M} \times {W_M} \times {C_M}}$; ${{\lambda _i}}$ is the scale weight.

\begin{equation}
\label{eq6}
{\cal L}{\rm{ = }}\sum\limits_{i = 0}^2 {{\lambda _i} \cdot F\left[ {f\left( {{{\bf{B}}_i}} \right),{\bf{\tilde M}}} \right]}  + {\lambda _3} \cdot F\left( {{\bf{M}},{\bf{\tilde M}}} \right)
\end{equation}

\section{Experiment}
\label{sec4}

\subsection{Implementation Details}
\label{sec4.1}
\subsubsection{Dataset}
\label{sec4.1.1}

The nuScenes {\color{blue}\citep{caesar2020nuscenes}} consists of 1k scenes from different cites, every scene contains 40 frames and lasts for the 20s, and every frame has 6 camera-view images with the direction toward front-left, front, front-right, back-left, back, and back-right directions. It provides the camera intrinsics and extrinsic matrixes for every frame and labels objects as 23 categories in the driving scene. Camera-view images are resized into ${224 \times 480}$ as described in {\color{blue}\citep{philion2020lift}}, and the 3d bounding boxes of objects are projected on the BEV plane to obtain the ground truth ${{\bf{\tilde M}}}$. Leveraging the vehicle as the original point in the BEV plane, there are two common size settings of ${{\bf{\tilde M}}}$, setting 1 {\color{blue}\citep{roddick2020predicting}} ranges from -25m to 25m on the horizontal axis and from -50m to 50m on the vertical axis, with a resolution of 0.25m, and setting 2 {\color{blue}\citep{zhou2022cross,philion2020lift,hu2021fiery}} ranges from -50m to 50m both on horizontal and vertical axes, with a resolution of 0.50m. We utilize the latter case as the primary size to demonstrate experimental results. Furthermore, following the category setting in {\color{blue}\citep{zhou2022cross, xu2023cobevt, chen2022efficient, philion2020lift}}, we will present the semantic segmentation result of vehicle and road.

The Argoverse {\color{blue}\citep{chang2019argoverse}} consists of 10k frame, and every frame has 7 camera-view images. We only utilize it in the generalization experiment with the setting in {\color{blue}\citep{zhou2022cross}}.

\subsubsection{Training}
\label{sec4.1.2}

We leverage the pre-trained EfficientNet-B4 {\color{blue}\citep{tan2019efficientnet}} as the encoder ${{\cal E}}$ to extract the features of camera-view images at layers 'reduction${\_}$4', 'reduction${\_}$3', 'reduction${\_}$2' as ${{{\bf{F}}_0}}$, ${{{\bf{F}}_1}}$, ${{{\bf{F}}_2}}$, and these camera-view features are downsampled with ${0.5 \times }$ to save computation. The sizes of ${{{\bf{B}}_0}}$, ${{{\bf{B}}_1}}$, ${{{\bf{B}}_2}}$ are set to 24, 48, and 100. All experiments were conducted on a single 3090 NVIDIA GPU by AdamW optimizer {\color{blue}\citep{loshchilov2017decoupled}} with the oncyclelr {\color{blue}\citep{smith2019super}} learning strategy, where the initial and maximum learning rates ${4{e^{ - 4}}}$and ${4{e^{ - 3}}}$. The iteration number is 70.4k with the batch size 8. Furthermore, other hyperparameters settings are shown in {\color{blue} Table \ref{table1}}.

\begin{table}[!ht]
\caption{The hyperparameters setting.}
\label{table1}
\centering
\begin{tabular}{cccccc}
\toprule           Hyperparameter & ${\xi }$	 & ${{\lambda _0}}$   &${{\lambda _1}}$   &${{\lambda _2}}$   &${{\lambda _3}}$  \\ \midrule
                          Value   &     0.05	 &1	                  &2	              & 2	              &60\\ \bottomrule
\end{tabular}
\end{table}

\subsubsection{Metric}
\label{sec4.1.3}
We assess the experimental results in terms of model accuracy and efficiency. Model accuracy is evaluated using Intersection over Union (IoU) for the vehicle and road categories, while model efficiency is evaluated in terms of the number of parameters (Paras) and Frames Per Second (FPS). To determine FPS, we gradually increase the batch size, with an exponent of 2, until the GPU utilization reaches 100\% without exceeding its memory capacity.

\subsection{Ablation Study}
\label{sec4.2}

In this section, we present the results of the vehicle in setting 2, unless otherwise specified.

\subsubsection{In Cross-Scale Hierarchical Transformer}
\label{sec4.2.1}
In this section, we verify the necessity of hierarchical framework, the scale setting in hierarchical framework, the necessity of cross-scale and residual connection, and comparison with other efficiency-oriented attentions.

\begin{figure}[!ht]
\centering
\includegraphics[width=3.5in]{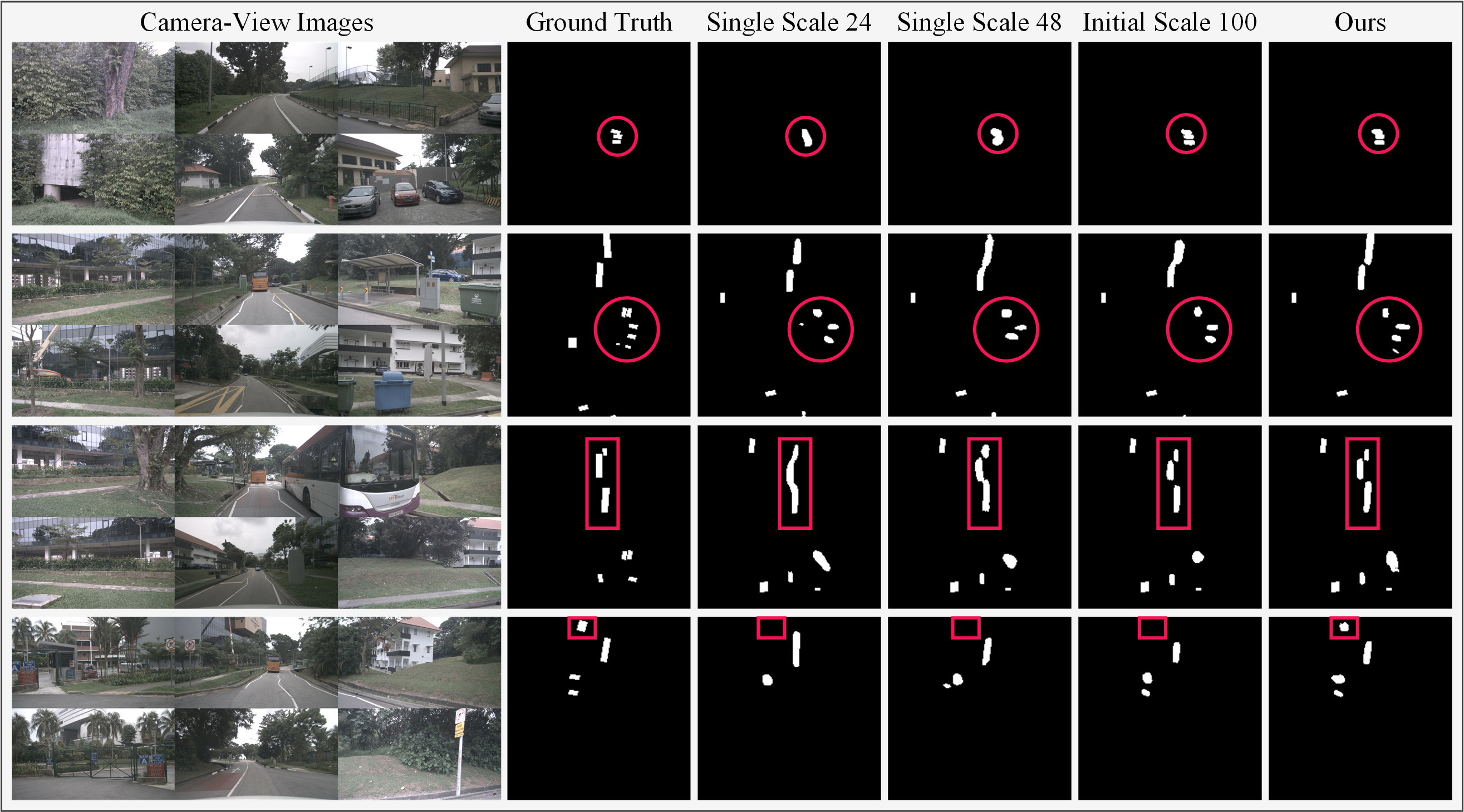}
\caption{The qualitative comparisons in verifying the necessity of hierarchical framework.}
\label{fig4}
\end{figure}

\textbf{The necessity of hierarchical framework} As described in {\color{blue}\citep{zhou2022cross}}, we set single scale cases of BEV learned feature as comparisons, where it learns the relationships with the multi-scale camera-view features and refines the last attention score to the final semantic segmentation via a convolution-based decoder. The single scale cases consist of 24, 48, and 100. Furthermore, we set the scale 100 as the initial scale and downsample it to obtain scales 48 and 24, where other experimental settings are the same as our method. This comparison case is to verify the rationality of refining the BEV feature from coarse to fine. The quantitative results are shown in {\color{blue} Table \ref{table2}}, and some qualitative instances are shown in {\color{blue} Fig. \ref{fig4}}.

\begin{table}[!ht]
\caption{The quantitative results of in verifying the necessity of hierarchical framework.}
\label{table2}
\centering
\begin{tabular}{cccc}
\toprule                     &IoU           &	Paras(M) &	FPS/batch  \\ \midrule
         Single Scale 24      &	35.305      &	5.060     &	486.2/32  \\
         Single Scale 48      &	35.202      &	5.625     &	511.5/16 \\
         Single Scale 100     &	-           &	9.069     &	200.1/4 \\
         Initial Scale 100    &	37.747      &	28.964     &	538.7/32 \\
                  Ours        &	38.682      &	7.762     &	431.6/32
\\ \bottomrule
\end{tabular}
\end{table}

For the case of single scale 24 and 48, their coarse-granularity feature representations are hard to capture the small objects in camera-view images and distinguish appressed instances. Consequently, they are prone to ignore objects far away from the vehicle in BEV semantic segmentation and infer the vehicle procession as one instance, and their IoUs reduce by approximately 3.43. For the case of single scale 100, its training cost numerous computations, where 24.67GB memory is occupied with batch size 8. Therefore, we only demonstrate its model efficiency, and their FPS is only 46.36\% of ours. As for the case of initial scale 100, the fined-granularity feature initialization disturbs the training stability, and this downsampling trick destroys the residual connection of BEV features reducing IoU 0.935.

\textbf{The scale setting in hierarchical framework} To verify the rationality of scale setting from 24 to 100, we set three scale setting cases as comparisons that are 12-24-50, 48-96-200, and 12-24-48-100. For the 12-24-48-100 case, we supply the camera-view feature at the 'reduction${\_}$1' layer to interact with the new BEV scale 12, and the purpose of this case is to verify the initial scale influence. The quantitative results are shown in {\color{blue} Table \ref{table3}}

\begin{table}[!ht]
\caption{The quantitative results in the scale setting.}
\label{table3}
\centering
\begin{tabular}{cccc}
\toprule                     &IoU           &	Paras(M) &	FPS/batch  \\ \midrule
               12-24-50      &	38.375      &	7.936     &	474.1/32  \\
           48-96-200         &	-           &	7.936     &	163.5/4 \\
         12-24-48-100        &	38.104      &	9.866     &	375.0/32 \\
                  Ours       &	38.682      &	7.762     &	431.6/32
\\ \bottomrule
\end{tabular}
\end{table}

The overall scales of BEV feature in the case of 12-24-50 are reduced by 50\%, and the coarser representations reduce IoU 0.307. In contrast, the case 48-96-200 doubles BEV feature scales but reduces FPS to only 163.5. Thus, our scale setting makes a better trade-off between accuracy and efficiency. The case 12-24-48-100 demonstrates that a coarser initial scale will not contribute to precision increase while cost more computation.

\textbf{The necessity of cross-scale and residual connection} To verify the effect of cross-scale Transformer in saving computation, we set the traditional scale-aligning Transformer as a comparison in model efficiency, namely, ${{{\bf{B}}_0}}$, ${{{\bf{B}}_1}}$, ${{{\bf{B}}_2}}$ learn the attention score with ${{{\bf{F}}_0}}$, ${{{\bf{F}}_1}}$, ${{{\bf{F}}_2}}$. Further, we pool ${{{\bf{F}}_0}}$, ${{{\bf{F}}_1}}$, ${{{\bf{F}}_2}}$ to ${0.125 \times }$ to enable the traditional scale-aligning Transformer to have the same model efficiency as ours, and to compare their model accuracy. Furthermore, to verify the function of residual connection in transmitting information between scales, as a comparison, we ablate the residual connection of ${{{\bf{B}}_i}}$ and only transfer the attention score ${{{\bf{A}}_i}}$ to the next scale. The quantitative results are shown in {\color{blue} Table \ref{table4}}..

\begin{table}[!ht]
\caption{The quantitative results of verifying the necessity of cross-scale and residual connection.}
\label{table4}
\centering
\begin{tabular}{cccc}
\toprule                         &IoU           &	Paras(M) &	FPS/batch  \\ \midrule
               Align w/o Pooling &	-           &	7.762     &	178.1/4  \\
           Align w/t Pooling     &	31.539      &	7.762     &	431.6/32 \\
         w/o Residual Connection &	38.304      &	7.433     &	484.9/32 \\
                  Ours           &	38.682      &	7.762     &	431.6/32
\\ \bottomrule
\end{tabular}
\end{table}

If we do not pool camera-view features, it demands 28.56GB memory to train the traditional scale-aligning Transformer, while only 15.00GB demanded for training our cross-scale Transformer. As a consequence, our experimental device is only capable of testing the traditional align-scale Transformer with batch size 4. By right of the parallel computing advantage for GPU, the FPS of cross-scale Transformer is 2.42 times as the traditional one. If we pool camera-view features, discarding details will impede the local relationship learning and the refinement of BEV feature representation. Furthermore, the case without residual connection weakens the information aggregation between scales and reduces IoU by 0.378.

\textbf{Comparison with other efficiency-oriented frames}
We compare the cross-scale hierarchical Transformer with other efficiency-oriented frames. For these comparisons, we select the same hierarchical structure in other frameworks but replace their attention blocks with local window attention {\color{blue}\citep{ liu2021swin}}, deformable attention {\color{blue}\citep{ zhu2020deformable}}, and channel attention {\color{blue}\citep{ ding2022davit}}. The quantitative results are shown in {\color{blue} Table \ref{table5}}.

\begin{table}[!ht]
\caption{The quantitative comparisons with other efficiency-oriented frames.}
\label{table5}
\centering
\begin{tabular}{cccc}
\toprule                             &IoU           &	Paras(M) &	FPS/batch  \\ \midrule
              Local Window Attention &	35.850      &	6.960     &	465.9/32  \\
           Deformable Attention      &	33.668      &	6.633     &	443.2/32  \\
         Channel Attention        &	30.916      &	7.934     &	468.9/32 \\
                  Ours               &	38.682      &	7.762     &	431.6/32
\\ \bottomrule
\end{tabular}
\end{table}

Local window attention allows each BEV token to absorb information from a specific region in camera-view images, leading to a decrease in IoU by 2.832. Though deformable attention retains global interaction, sparse relationship learning exhibits a more significant decrease in IoU. On the other hand, Channel attention sacrifices the spatial relationship learning resulting in BEV tokens being hindered in their adaptive exploration of correspondences in space, with an IoU of only 30.916. In contrast, our cross-scale hierarchical frame achieves a favorable trade-off between efficiency and effectiveness.

\subsubsection{In Correspondence-Augmented Attention}
\label{sec4.2.2}
In this section, we verify the necessity of correspondence augmentation, augmentation dimension, and augmentation degree. To evaluate the correspondence degree, we endeavor to analyze the distribution of the position-view-related and position-view-disrelated attention scores. As described in {\color{blue} Sec. \ref{sec3.3.1}}, we regard ${{{\bf{M}}_u} - \left\{ {{{\bf{I}}_{fl}},{{\bf{I}}_f},{{\bf{I}}_{fr}}} \right\}}$ and ${{{\bf{M}}_l} - \left\{ {{{\bf{I}}_{bl}},{{\bf{I}}_b},{{\bf{I}}_{br}}} \right\}}$ as conducive correspondences and treat ${{{\bf{M}}_l} - \left\{ {{{\bf{I}}_{fl}},{{\bf{I}}_f},{{\bf{I}}_{fr}}} \right\}}$ and ${{{\bf{M}}_u} - \left\{ {{{\bf{I}}_{bl}},{{\bf{I}}_b},{{\bf{I}}_{br}}} \right\}}$ as inconducive correspondences. We count the attention score quantity of the conducive and inconducive correspondences in the range of ${\left[ {0.7,1.0} \right]}$ and ${\left[ {0.1,1.0} \right]}$.

\begin{figure}[!ht]
\centering
\includegraphics[width=3in]{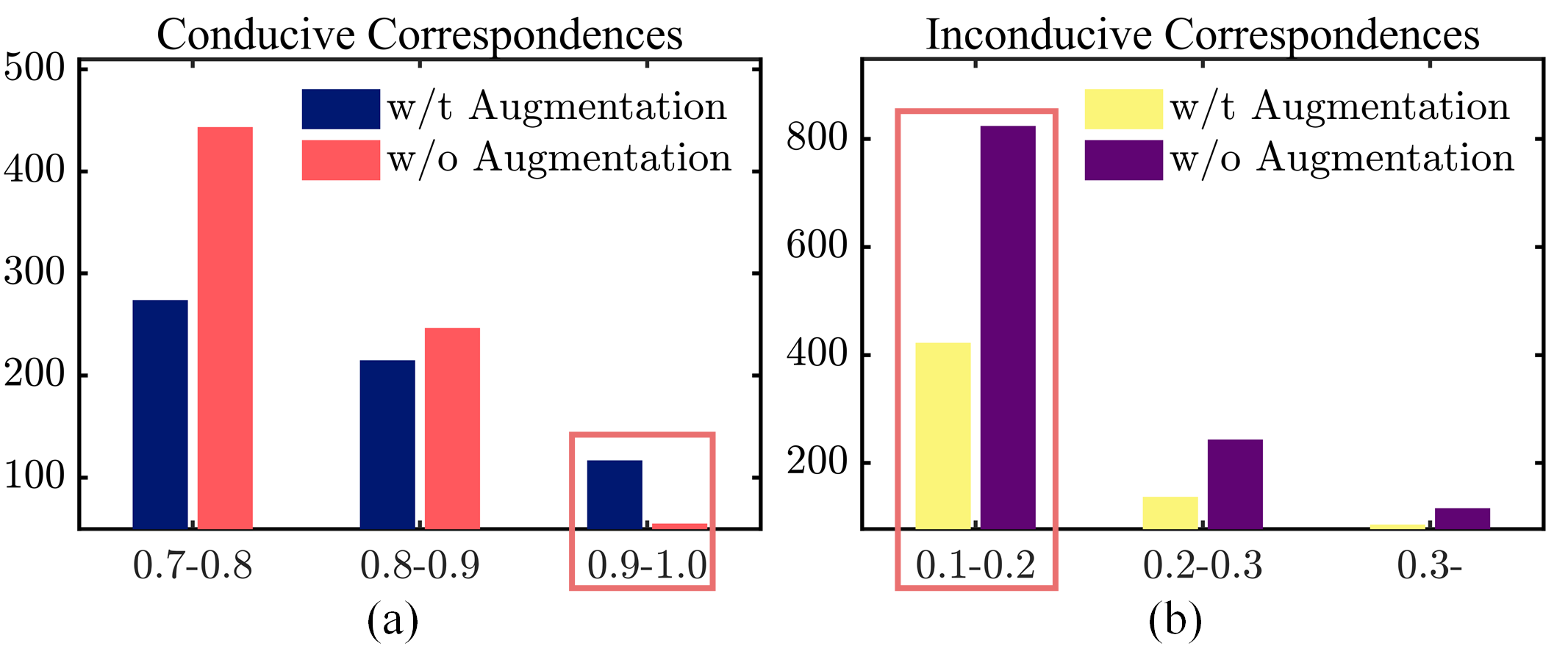}
\caption{(a) The attention score distribution of the conducive correspondence; (b) The attention score distribution of the inconducive correspondence. The ${x}$-axis depicts the attention scores range, while the ${y}$-axis depicts the number of attention scores in a specific range.}
\label{fig5}
\end{figure}

\textbf{The necessity of correspondence augmentation} We ablate the correspondence-augmented attention as a comparison, namely, do not amplify attention scores before the Softmax normalization. The analysis of conducive and inconducive correspondences are shown in {\color{blue} Fig. \ref{fig5}}, and the precision and efficiency comparisons are shown in {\color{blue} Table \ref{table6}}.

\begin{table}[!ht]
\caption{The quantitative results of verifying the necessity of correspondence augmentation.}
\label{table6}
\centering
\begin{tabular}{cccc}
\toprule                         &IoU           &	Paras(M) &	FPS/batch  \\ \midrule
               w/o Correspondence Augmentation  &	38.358    &7.762     &	433.8/32  \\
                  w/t Correspondence Augmentation&	38.682    &7.762     &	431.6/32
\\ \bottomrule
\end{tabular}
\end{table}

The correspondence-augmented attention increases IoU 0.324, without causing a conspicuous FPS decrease. As {\color{blue} Fig. \ref{fig5}} shows, we endeavor to analyze the IoU increase reason in the attention score distribution aspect. If an attention score between one camera-view token and one BEV token is over 0.9, this camera-view token will contribute major information in representing this BEV token. As the correspondence-augmented attention doubles the attention score quantity in the range of ${\left[ {0.9,1.0} \right]}$, it demonstrates that the correspondence-augmented attention is prone to highlight the dominant relationship and increases its attention score. The case with correspondence augmentation still has more quantity in the range greater than 0.8 but has less quantity in the range of ${\left[ {0.7,0.8} \right]}$. It demonstrates that correspondence-augmented attention enables the overall position-view attention score to approach 1.0. Note that, the attention score of inconducive correspondence greater than 0.1 will have a significant negative impact on the fusion of conducive correspondence, the correspondence-augmented attention enables the attention score quantity in this range to fall approximately in half, and it reveals that it is capable of suppressing absorbing information from inconducive correspondence.

\begin{figure}[!ht]
\centering
\includegraphics[width=2.6in]{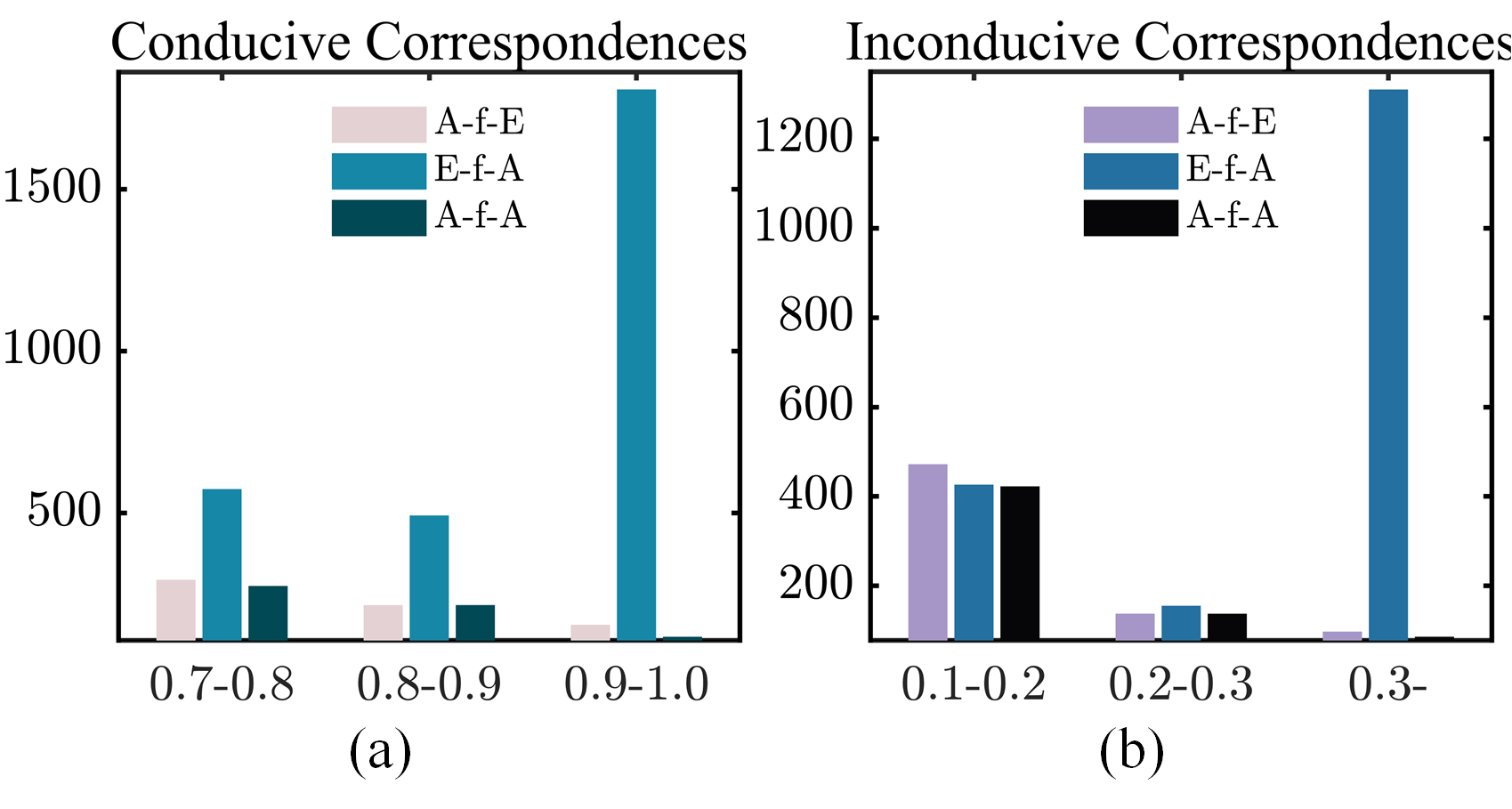}
\caption{(a) The attention score distribution of the conducive correspondence; (b) The attention score distribution of the inconducive correspondence. The ${x}$-axis and ${y}$-axis are the same as in Fig. 5. The A-f-E, E-f-A, and A-f-A are related to the three cases of TABLE VI.}
\label{fig6}
\end{figure}

\textbf{Augmentation dimension} As described in {\color{blue} Sec. \ref{sec3.2.2}}, for every BEV token, it calculates attention scores with all camera-view tokens for all views, and the local amplification coefficients are calculated based on all attention scores. As comparisons, we calculate local amplification coefficients based on all camera-view tokens for every view and every camera-view token for all views. In this way, we verify the rationality of the augmentation dimension. The analysis of their conducive and inconducive correspondences are shown in {\color{blue} Fig. \ref{fig6}}, and the precision and efficiency comparisons are shown in {\color{blue} Table \ref{table7}}.

\begin{table}[!ht]
\caption{The quantitative results for various augmentation dimension.}
\label{table7}
\centering
\begin{tabular}{cccc}
\toprule                         &IoU           &	Paras(M) &	FPS/batch  \\ \midrule
               All Camera-View Tokens for Every View  &	37.761    &7.762     &	383.5/32  \\
               Every Camera-View Token for All Views  &	38.239    &7.762     &	393.9/32 \\
               All Camera-View Tokens for All Views   &	38.682    &7.762     &	431.6/32

\\ \bottomrule
\end{tabular}
\end{table}

Calculating the local amplification coefficient based on all camera-view tokens for every view means to distinguishing the conducive and inconducive correspondence in a single view. However, as described in {\color{blue} Sec. \ref{sec3.3.1}}, all correspondences for a view maybe are conducive or inconducive. Thus, this correspondence augmentation method enlarges some attention scores of inconducive correspondence, namely, the inconducive correspondence is highlighted rather than suppressed. As {\color{blue} Fig. \ref{fig6}(b)} shows, the number of its attention score greater than 0.1 increases by 9.46\%, and its IoU decreases by 0.921. Since the view quantity is much smaller than the camera-view token quantity, in calculating the local amplification coefficient based on every camera-view token for all views, the standard deviation is random and varies greatly. As a consequence, more correspondences are considered as conducive and are highlighted, both conducive and inconducive correspondences leap, and it decreases IoU by 0.444. In contrast, our correspondence augmentation method conforms to the intention in {\color{blue} Sec. \ref{sec3.3.1}} and has the best IoU among them.

\begin{figure}[!ht]
\centering
\includegraphics[width=3in]{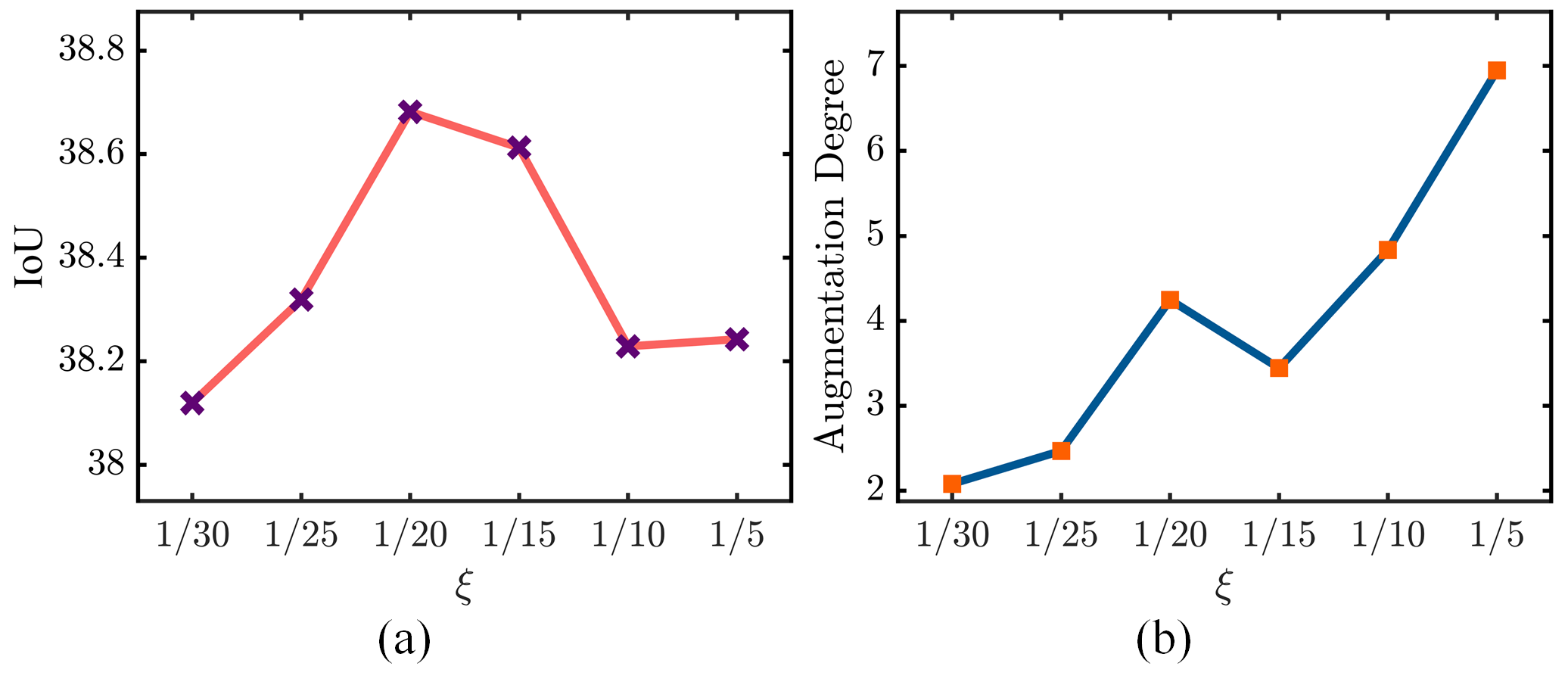}
\caption{(a) The IoU variation with ${\xi }$; (b) The augmentation degree variation with ${\xi }$. The augmentation degree is ${{\sigma _i}}$ multiplying with ${\xi }$.}
\label{fig7}
\end{figure}

\textbf{Augmentation degree} To verify the augmentation degree impact on model precision, we set the parameter ${\xi }$ to ${{1 \mathord{\left/
 {\vphantom {1 {30}}} \right.\kern-\nulldelimiterspace} {30}}}$, ${{1 \mathord{\left/{\vphantom {1 {25}}} \right.\kern-\nulldelimiterspace} {25}}}$, ${{1 \mathord{\left/{\vphantom {1 {15}}} \right.
\kern-\nulldelimiterspace} {15}}}$, ${{1 \mathord{\left/{\vphantom {1 {10}}} \right.\kern-\nulldelimiterspace} {10}}}$, and ${{1 \mathord{\left/{\vphantom {1 {15}}} \right.\kern-\nulldelimiterspace} {15}}}$ as comparisons. The quantitative results are shown in {\color{blue} Fig. \ref{fig7}}.

When ${\xi }$ is 1/30 or 1/25, the augmentation degree is approximately 2, their inferring precision is close to the case without correspondence augmentation, and it reveals these cases are incapable of highlighting conducive correspondence to a certain extent. When ${\xi }$ is lower than 1/30, a worse case appears that the attention scores of conducive and inconducive correspondence distribute towards middle. When ${\xi }$ is 1/10 or 1/5, the augmentation degree is around 6, and the augmentations in these cases work as finding a maximum value so that some conducive correspondences also are suppressed. As a consequence, Transformer only absorbs information from the most related camera-view token in representing a BEV token. However, an object may appear both in ${{{\bf{I}}_f}}$ and ${{{\bf{I}}_{fl}}}$, so the correspondence augmentations for aforementioned cases impede the effective information fusion between multi-views and reduces the IoU score. In contrast, the augmentation degree approach 4 when ${\xi }$ is 1/20 or 1/15, which ensures the highlighting and suppressing for attention scores are moderate.

\textbf{Comparison with other augmentation strategies} We compare the correspondence augmentation with other augmentation strategies in terms of channel and view. Specifically, we employ a SENet {\color{blue}\citep{ hu2018squeeze}} behind each ${{{\bf{{ A}}}_i}}$ to augment channels. For view augmentation, we exploit an additional full attention block to learn relationships between a specific view and the remaining views and add these learned relationships into camera-view image features. The quantitative results are shown in {\color{blue} Table \ref{table8}}.

\begin{table}[!ht]
\caption{The quantitative comparisons with other augmentation.}
\label{table8}
\centering
\begin{tabular}{cccc}
\toprule                               &    IoU           &	Paras(M)  &	FPS/batch  \\ \midrule
              w/t Channel Augmentation &	36.290        &	8.656     &	286.1/32  \\
           w/t View Augmentation       &	37.889        &	11.620    &	27.8/16  \\
         Ours                           &	38.682        &	7.762     &	431.6/32
\\ \bottomrule
\end{tabular}
\end{table}

Both channel and view augmentation fail to improve inference precision while costing more computation. In contrast, our correspondence augmentation inspires by the insight of information absorption and aggregation, redistributing attention scores to achieve a precision increase without introducing additional modules.

\subsubsection{In objective function}
\label{sec4.2.3}

In this section, we endeavor to verify the function of supervising multi-scale BEV representations in our objective function. As a comparison, we only supervise the final semantic segmentation with ground truth, and the quantitative results are shown in {\color{blue} Table \ref{table9}}.

\begin{table}[!ht]
\caption{The quantitative comparison in verifying objective function.}
\label{table9}
\centering
\begin{tabular}{cccc}
\toprule                         &IoU           &	Paras(M) &	FPS/batch  \\ \midrule
               Only Supervising ${{\bf{M}}}$    &	38.623    &6.874     &	642.1/32  \\
               Full Objective Function          &	38.682    &7.762     &	393.9/32

\\ \bottomrule
\end{tabular}
\end{table}

\begin{figure}[!ht]
\centering
\includegraphics[width=1.8in]{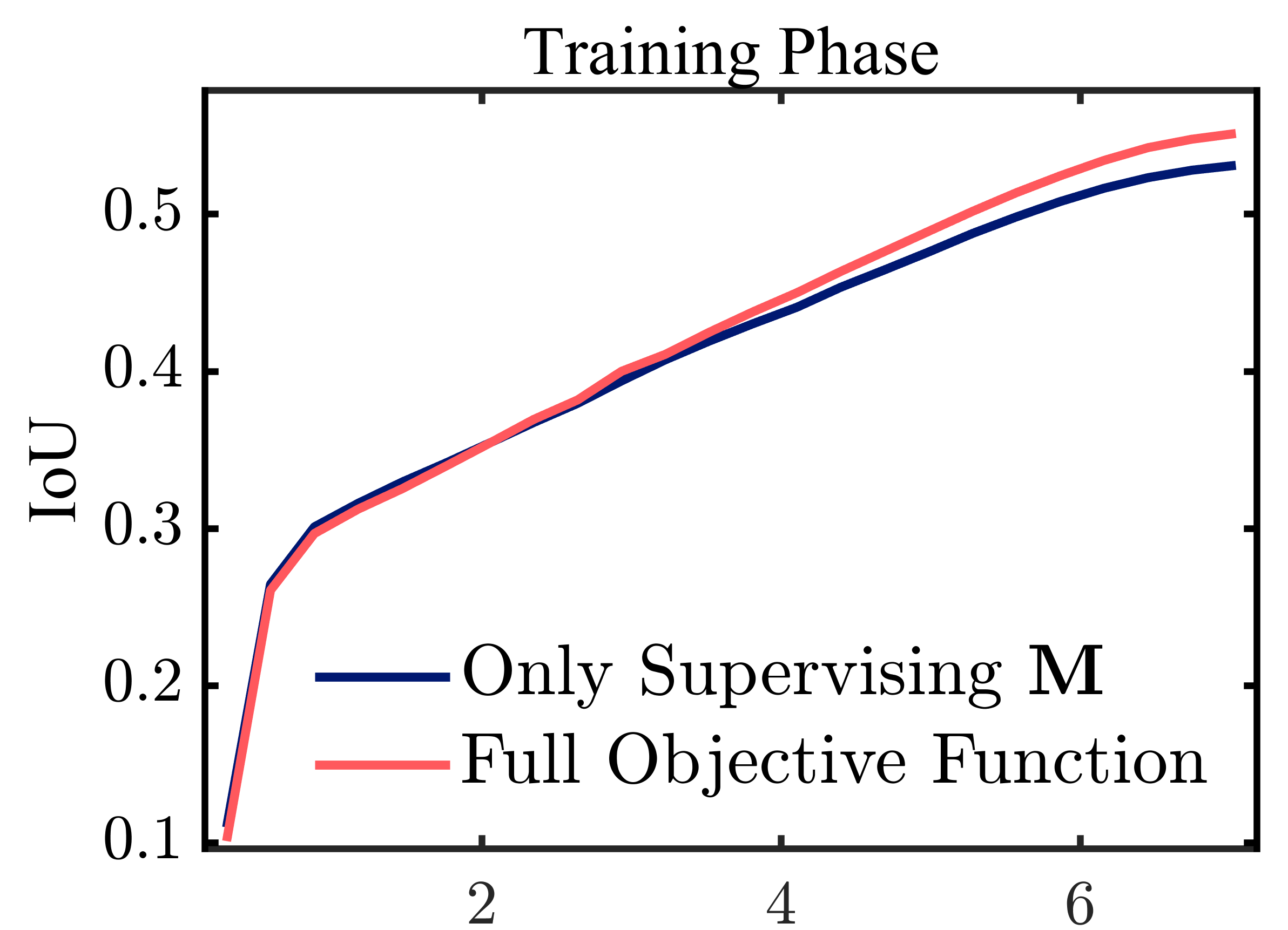}
\caption{The IoU variation tendency comparisons in the training phase.}
\label{fig8}
\end{figure}

Supervising the multi-scale BEV representations in the training phase increase 0.059 IoU, and this step-by-step supervising method facilitates the model to coverage quickly shown in {\color{blue} Fig. \ref{fig8}}. As the layer combination ${f\left(  \cdot  \right)}$ described in {\color{blue} Sec. \ref{sec3.4}}, FPS decreases with the parameter number increasing. However, these branches ${f\left(  \cdot  \right)}$ have no function in the inference phase, if we discard them in the inference phase, there will be no FPS loss.

\subsubsection{The robustness of camera dropping}
\label{sec4.2.4}

\begin{figure}[!ht]
\centering
\includegraphics[width=1.8in]{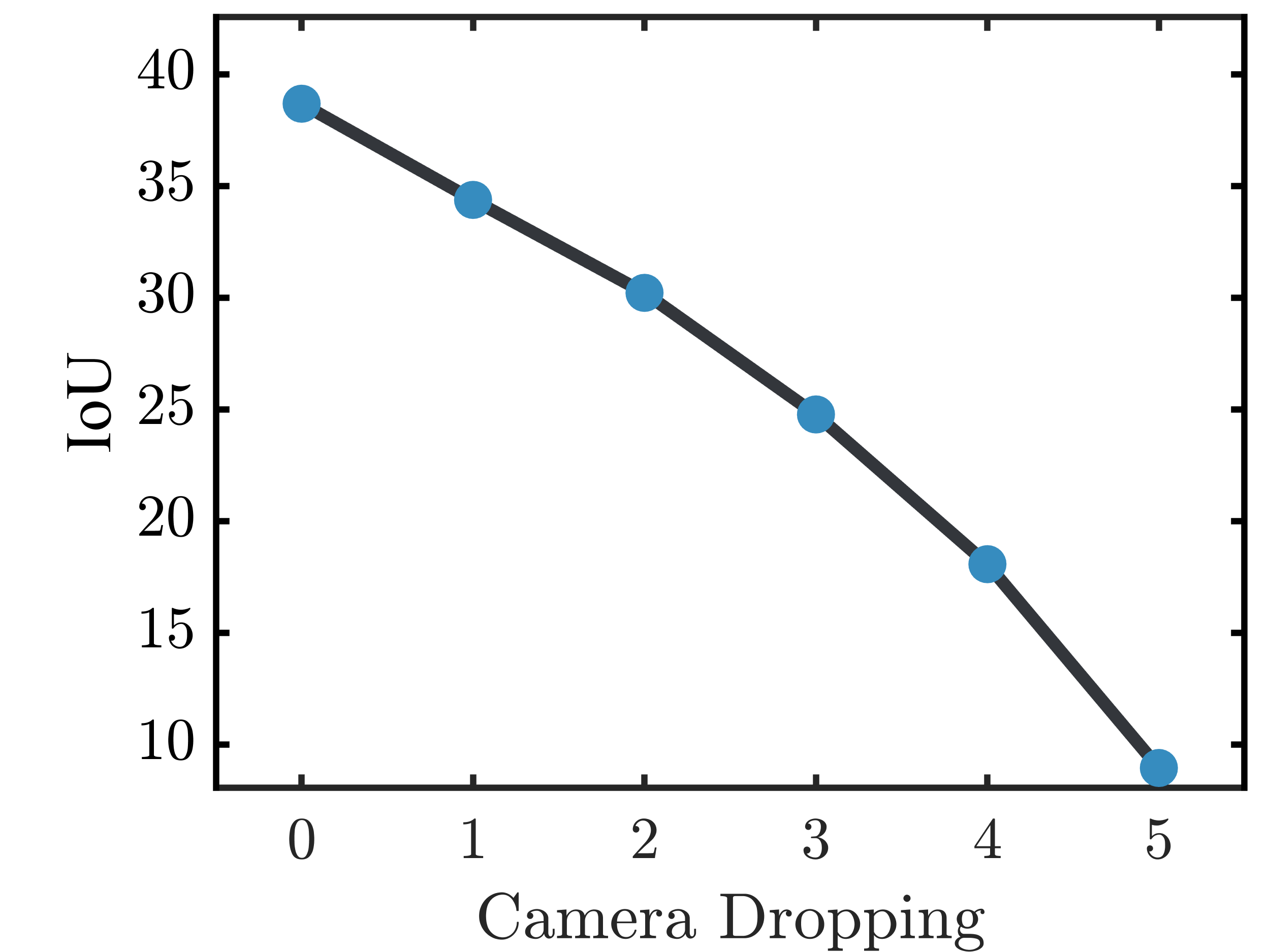}
\caption{The IoU variation tendency with camera dropping.}
\label{fig9}
\end{figure}

As the number of cameras is set to 6 as default, to assess our method performance in other camera quantity cases, we devise the robustness experiments of camera dropping, the camera-view images of some views are discarded in the training and testing phase. The model precision variation is shown in {\color{blue} Fig. \ref{fig9}}.

When the camera dropping quantity is 1 or 2, the remaining cameras still capture most surrounding scenes, IoU is still greater than 30 and decreases slowly. Since the correspondence-augmented attention is capable of highlighting the conducive one, it slows down IoU decrease for the above case to some extent. When the camera dropping quantity is 4 or 5, the circumstance is similar to inferring ${360^\circ }$ surrounding scene conditioned on a monocular image, so that IoU decreases quickly. In a nutshell, our method is robust to emergency circumstances with little camera dropping.

\subsection{Comparison}
\label{sec4.3}

\begin{figure*}[!ht]
\centering
\includegraphics[width=6in]{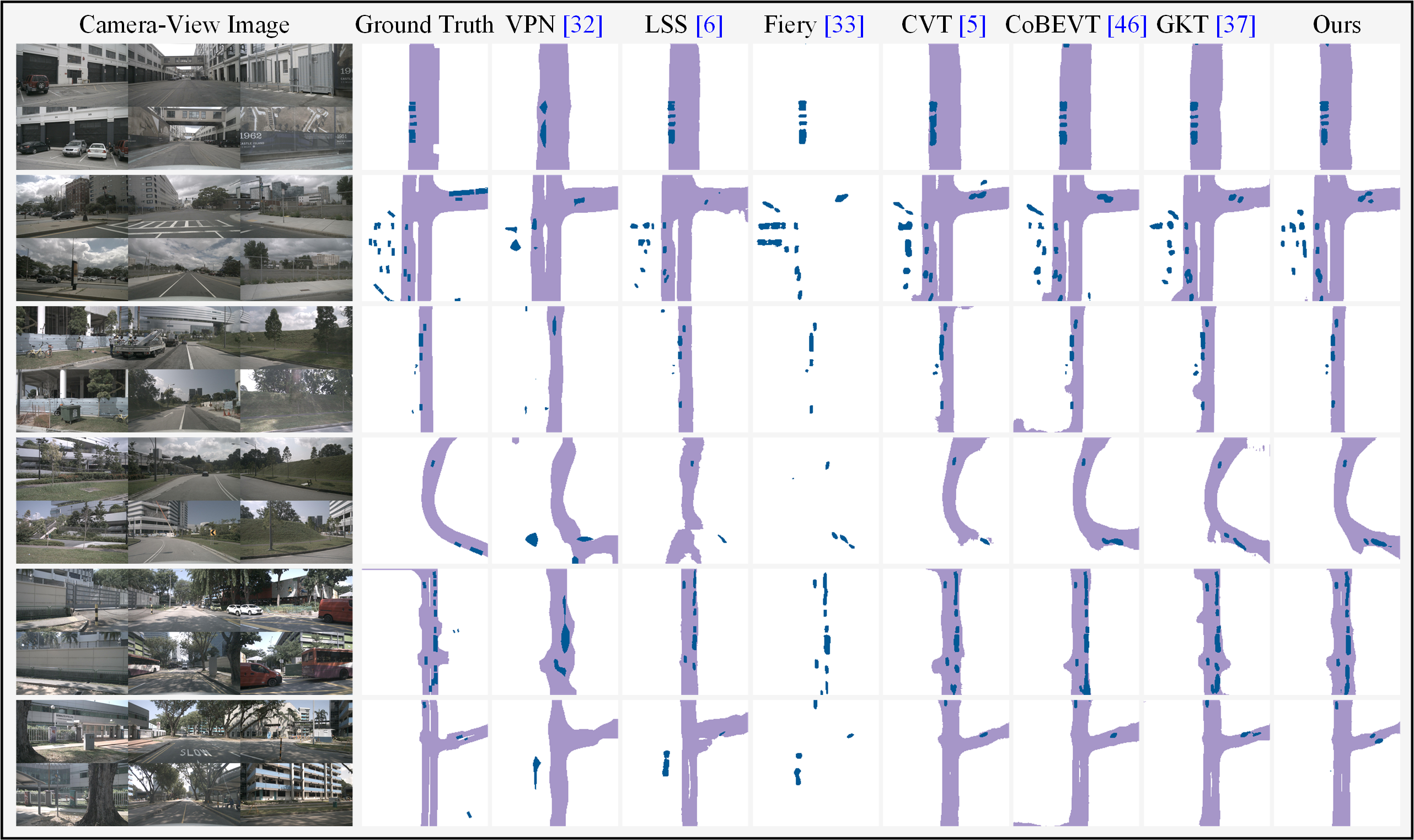}
\caption{Some examples of method comparison. We leverage the first three and last three examples to indicate the superiority of our method in representing fine-grained adjoining instances and far objects. Since Fiery only released its implementation version of vehicle semantic segmentation, we do not display its road part in comparisons.}
\label{fig10}
\end{figure*}

For a fair comparison, we choose the method conditioned on multi-camera-view images as baselines. We compare with VPN {\color{blue}\citep{pan2020cross}}, LSS {\color{blue}\citep{philion2020lift}}, Fiery {\color{blue}\citep{hu2021fiery}}, CVT {\color{blue}\citep{zhou2022cross}}, CoBEVT {\color{blue}\citep{xu2023cobevt}} and GKT {\color{blue}\citep{chen2022efficient}}. We directly trained VPN on the nuScenes dataset leveraging generated ground truth instead of following their domain transfer method, and CVT, CoBEVT, GKT were trained with the same iteration as ours for better performances. The quantitative results are shown in {\color{blue} Table \ref{table10}}, and some qualitative comparisons are shown in {\color{blue} Fig. \ref{fig10}}.

\begin{table*}[!ht]
\caption{The quantitative comparison between different methods. \textbf{Bold} and \underline{underline} highlights represent the best and second-best performance.}
\centering
\label{table10}
\begin{tabular}{@{}lllllllll@{}}
\toprule
                  & \multicolumn{4}{c}{Setting 2}                    & \multicolumn{4}{c}{Setting 1}                   \\ \cmidrule(l){2-9}
                                                    & IoU(vehicle)    & IoU(road) & Paras(M) & FPS/batch  & IoU(vehicle) & IoU(road) & Paras(M) & FPS/batch \\ \midrule
VPN     {\color{blue}\citep{pan2020cross}}          & 13.158            & 63.938            & 14.413   & 743.3/32   & 24.002            & 73.137        & 15.339   & 254.0/32  \\
LSS     {\color{blue}\citep{philion2020lift}}       & 32.039            & 60.755            & 9.043    & 62.2/64    & 33.184            & 60.786        & 9.043    & 56.2/32         \\
Fiery   {\color{blue}\citep{hu2021fiery}}           & 35.766            & -                 & 3.591    & 6.7/32     & -                 & -             & -        & -                \\
CVT     {\color{blue}\citep{zhou2022cross}}         & 37.648            & \underline{72.152}& 4.321    & 627.9/16   & \underline{40.935}& 78.239        & 4.641    & 298.1/8          \\
CoBEVT  {\color{blue}\citep{xu2023cobevt}}          & 37.704            & 70.635            & 6.439    & 77.7/32    & 39.513            & 77.370        & 7.719    & 195.5/16         \\
GKT     {\color{blue}\citep{chen2022efficient}}     & \underline{37.818}& 71.606            & 4.701    & 466.0/32   & \textbf{41.682}   & \textbf{78.944}& 5.021   & 211.1/8          \\
Ours                                                & \textbf{38.682}   & \textbf{72.536}   & 7.762    & 431.6/32   & 40.817            & \underline{78.425}& 8.057 & 330.5/16         \\ \bottomrule
\end{tabular}
\end{table*}

As the first three examples show, benefiting from our hierarchical framework and BEV representation from coarse to fine, our method is capable of distinguishing various instances in the adjoining vehicle procession. In contrast, CVT, CoBEVT, and GKT represent fine-grained instances poorly since it learns relationships in the latent layer. As the last three examples show, since the correspondence-augmented attention is capable of highlighting the position-view-related attention score, our method is expert in capturing and representing objects far from the center vehicle, which will contribute to the early emergency warning and win more time for downstream tasks. In general, for setting 2, our method has state-of-the-art performance in inferring both vehicle and road semantic segmentations. As described in {\color{blue} Sec. \ref{sec4.2.3}}, our method is capable of reaching 642.1 FPS, which is top-ranking in comparison. As setting 1 gives up to infer some far objects, the superiority of our method is weakened but still has a high performance.

\subsection{Generalization}
\label{sec4.4}
We further compare with other methods on the Argoverse dataset. As shown in {\color{blue} Table \ref{table11}},  our method still has a comparable performance in vehicle and road semantic segmentation tasks. In addition, we adapt our method from the binary classification task into the multi-classification task by regularizing the projection head, and the semantic segmentation results for the original categories on the nuScenes dataset are shown in {\color{blue} Table \ref{table12}}.

\begin{table}[!ht]
\caption{The quantitative comparison on the Argoverse dataset. \textbf{Bold} and \underline{underline} highlights represent the best and second-best performance.}
\centering
\label{table11}
\begin{tabular}{@{}lll@{}}
\toprule
                                            & IoU(vehicle) & IoU(road) \\ \midrule
LSS {\color{blue}\citep{philion2020lift}}   & 20.498            & 54.891                    \\
CVT {\color{blue}\citep{zhou2022cross}}     & 29.990            & 59.673                    \\
CoBEVT {\color{blue}\citep{xu2023cobevt}}   & \underline{32.429}& \textbf{61.432}           \\
Ours                                        & \textbf{33.878}   & \underline{60.056}         \\ \bottomrule
\end{tabular}
\end{table}

\begin{table}[!ht]
\caption{The quantitative results of our method on the nuScenes dataset for original categories.}
\centering
\label{table12}
\begin{tabular}{@{}llll@{}}
\toprule
             & IoU &               & IoU \\ \midrule
Car          & 39.626   & Motorcycle    & 1.570   \\
Truck        & 20.756   & Bicycle       & 1.527   \\
Bus          & 12.022   & Lane          & 53.619   \\
Trailer      & 10.372   & Road\_Segment & 70.481   \\
Construction & 1.893    &              &     \\ \bottomrule
\end{tabular}
\end{table}

\section{Conclusion}
\label{sec5}
This paper proposes a novel cross-scale hierarchical Transformer with correspondence-augmented attention for inferring BEV semantic segmentation method. We devise a hierarchical frame to refine the representation of BEV features, which is initialized at a coarse scale and learns relationships with multi-scale features of camera-view images until it reaches the half size of the final segmentation. To circumvent the computation cost leap caused by this hierarchical frame, the relationship learning is implemented in a reversed-aligning way for the cross-scale Transformer, and we exploit the residual connection of BEV features to compensate for the potential accuracy loss. As not all camera-view tokens contribute to representing a specific BEV token, we propose correspondence-augmented attention to distinguish conducive and inconducive correspondences based on the view and BEV token position. Standard deviation is calculated for every BEV token, and attention scores are amplified accordingly. Later, the Softmax operation will distribute attention scores toward extreme large and small, highlighting and suppressing the position-view-related and the position-view-disrelated attention scores. The proposed method achieves state-of-the-art performance in inferring BEV semantic segmentation conditioned on multi-camera-view images, with 38.682 and 72.536 IoU for vehicle and road categories on the nuScenes dataset, respectively.

Our cross-scale hierarchical framework reduces the computation cost at the point of block combination. However, each block still employs full attention, implying that it possesses the potential to enhance inference efficiency from the perspective of block mechanism. In our future work, we will focus on an efficiency-oriented and correspondence-augmented attention mechanism, with the goal of controlling attention sparsity to reinforce the relationship between BEV tokens and views while also reducing computation cost.

\bibliographystyle{IEEEtran}
\bibliography{bare_jrnl_new_sample4}{}

\begin{IEEEbiography}[{\includegraphics[width=1in,height=1.25in,clip,keepaspectratio]{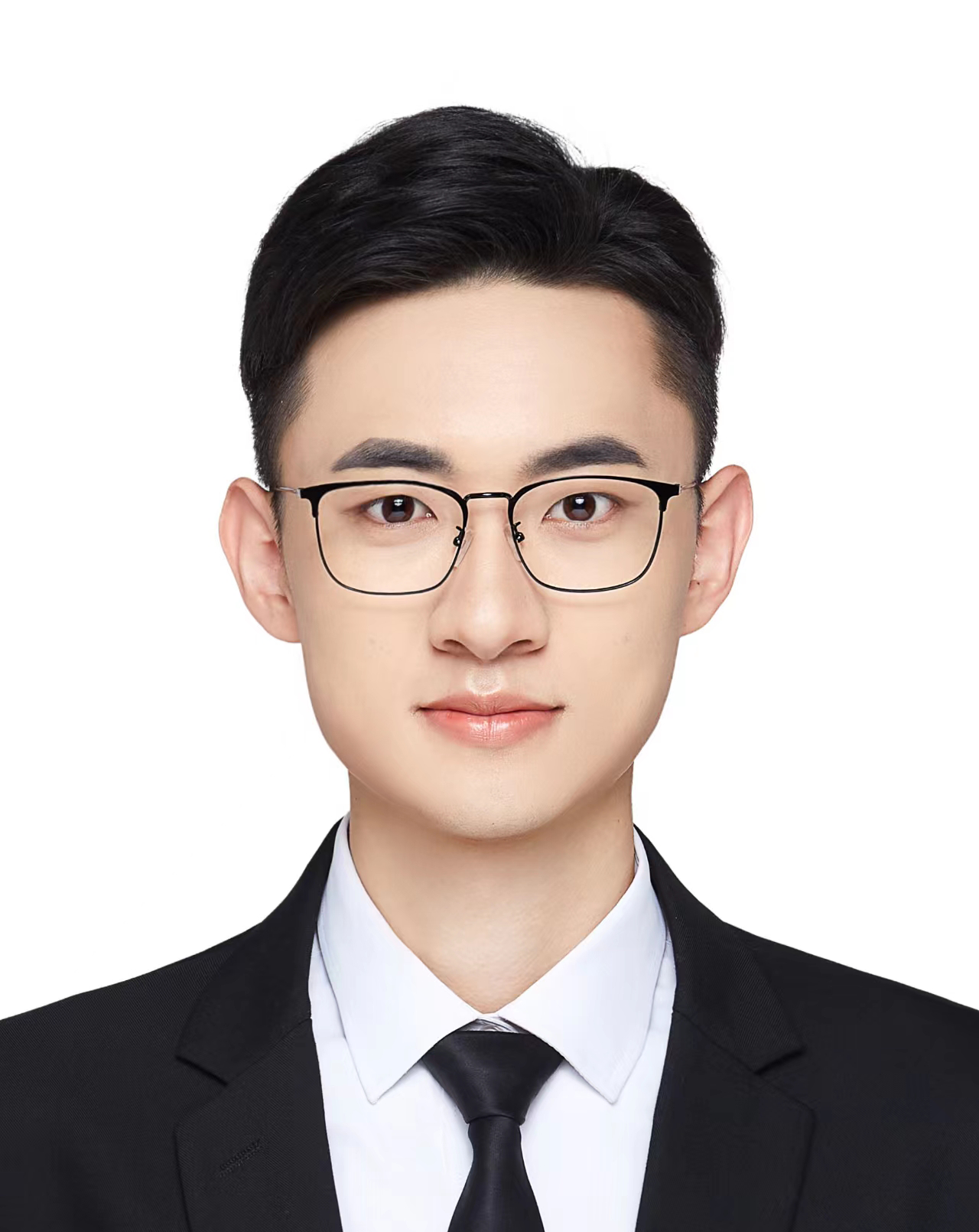}}]{Naiyu Fang}
received the B.Eng. degree in the School of Mechanical Engineering from Dalian University of Technology, Dalian, China, in 2019. He is currently a Ph.D. candidate in the School of Mechanical Engineering, Zhejiang University, China. His research interests include autonomous driving and virtual try-on.\end{IEEEbiography}

\begin{IEEEbiography}[{\includegraphics[width=1in,height=1.25in,clip,keepaspectratio]{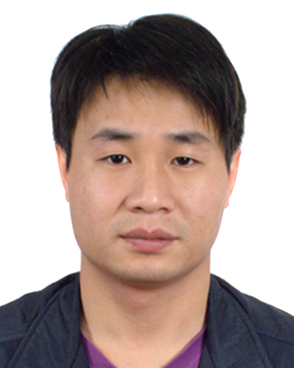}}]{Lemiao Qiu}
received the Ph.D. degree from the Department of Mechanical Engineering, Zhejiang University, Hangzhou, China, in 2008. He is currently an Associate Professor at the Department of Mechanical Engineering, Zhejiang University, China. His research interests include computer graphics, autonomous driving, and production informatization.\end{IEEEbiography}

\begin{IEEEbiography}[{\includegraphics[width=1in,height=1.25in,clip,keepaspectratio]{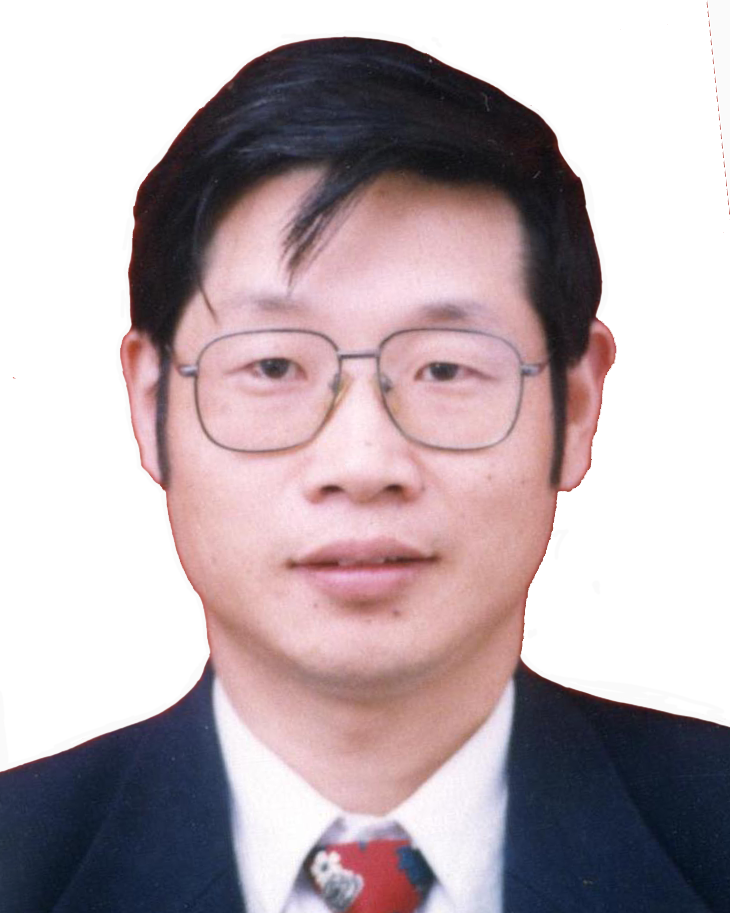}}]{Shuyou Zhang}
is currently a Distinguished Professor and a Ph.D. Supervisor at the Department of Mechanical Engineering, Zhejiang University, China. His research interests include computer graphics, computer vision, and product digital design.\end{IEEEbiography}

\begin{IEEEbiography}[{\includegraphics[width=1in,height=1.25in,clip,keepaspectratio]{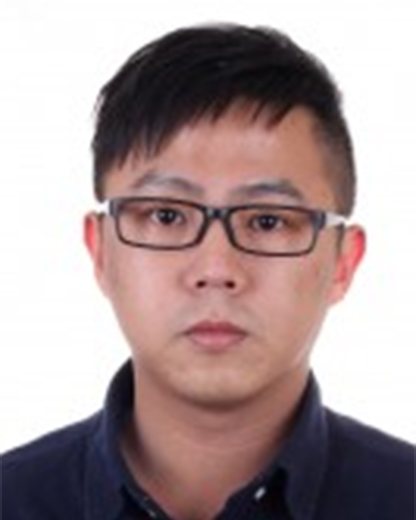}}]{Zili Wang}
received the Ph.D. degree from the Department of Mechanical Engineering, Zhejiang University, Hangzhou, China in 2018. He is currently Research Associate at the Department of Mechanical Engineering, Zhejiang University, Hangzhou, China. His research interests include computer-aided design and computer graphics.\end{IEEEbiography}

\begin{IEEEbiography}[{\includegraphics[width=1in,height=1.25in,clip,keepaspectratio]{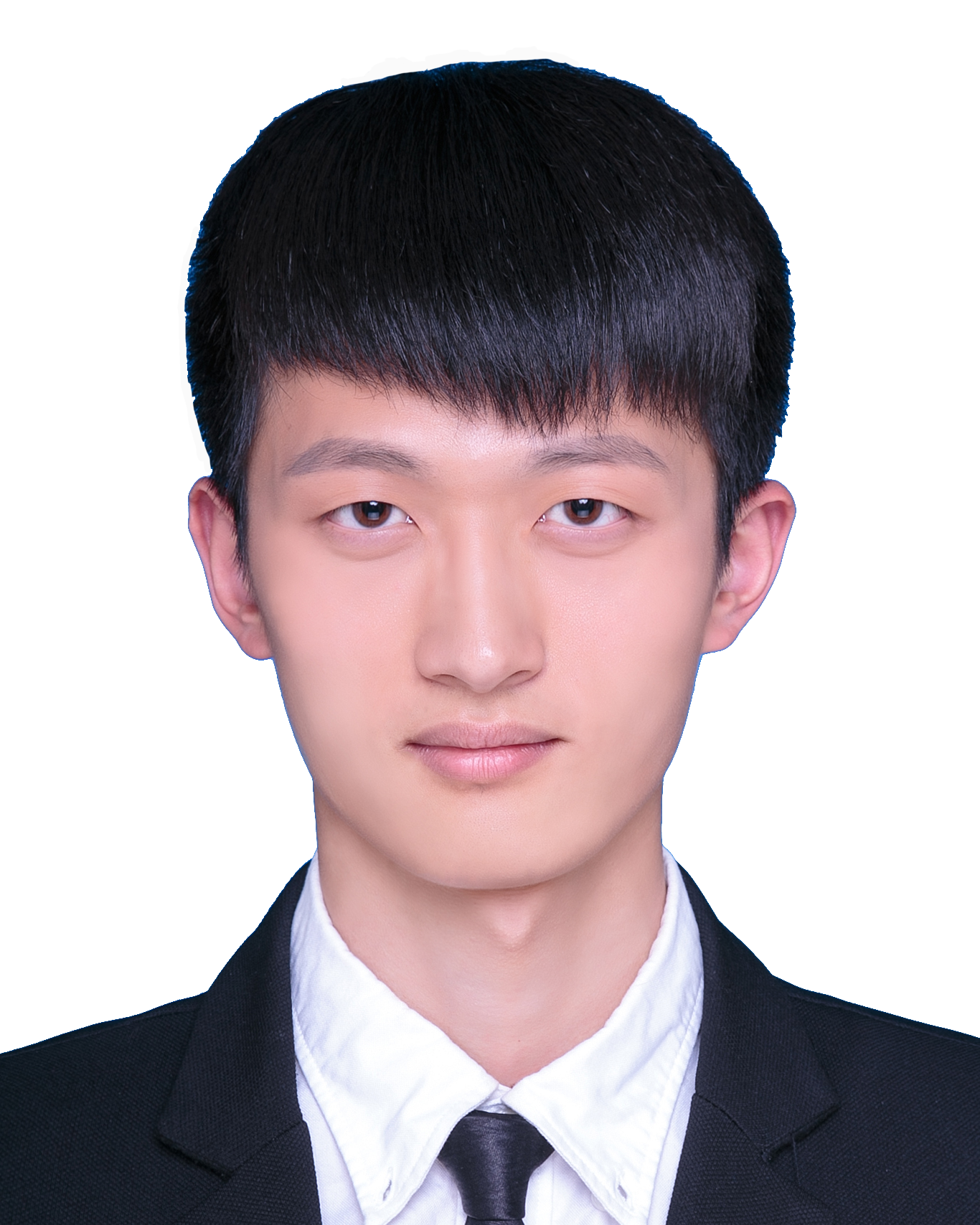}}]{Kerui Hu}
received the B.Eng. degree in the School of China University of Mining and Technology, Xuzhou, China, in 2018. He is currently a Ph.D. candidate in the School of Mechanical Engineering, Zhejiang University, China. His research interests include data mining and collaborative filtering.
\end{IEEEbiography}

\begin{IEEEbiography}[{\includegraphics[width=1in,height=1.25in,clip,keepaspectratio]{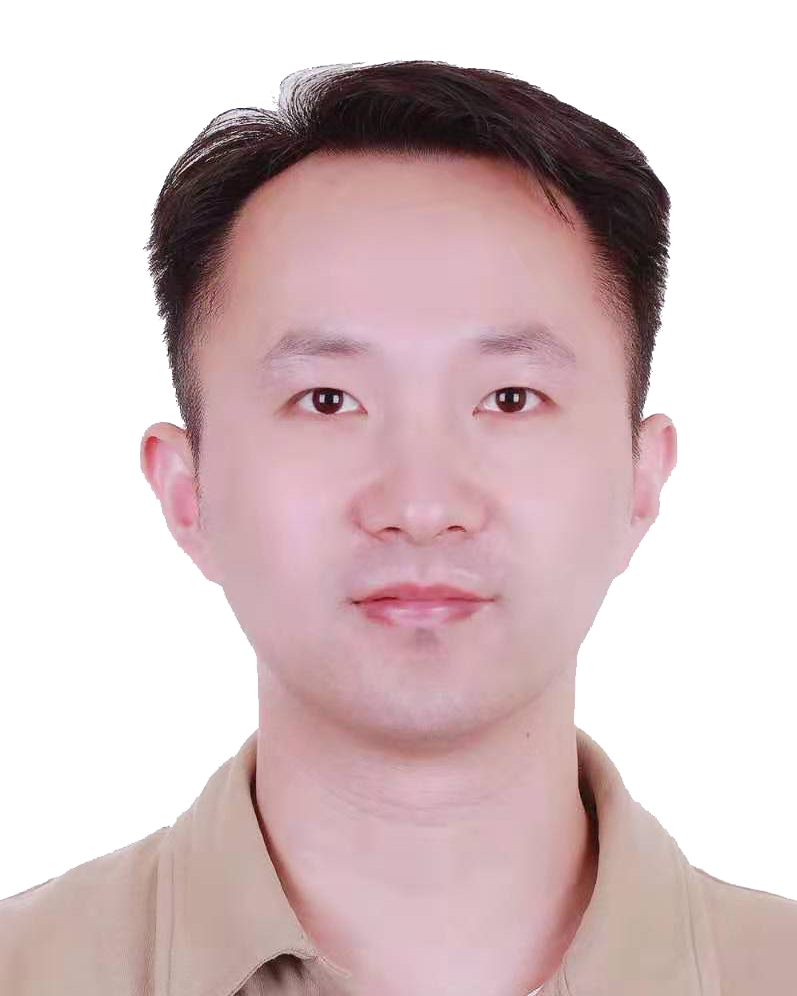}}]{Kang Wang}
received the PhD degree from the Department of Mechanical Engineering, Zhejiang University, Hangzhou, China in 2021. He is currently a post-doc researcher in the Hong Kong Polytechnic University, Hong
Kong. His research interests include deep learning and computer vision.
\end{IEEEbiography}

\vfill

\end{document}